\documentclass{article}

    \usepackage[preprint]{neurips_2023}

\usepackage{amsmath}
\usepackage{amssymb}
\usepackage{bbm}
\usepackage[ruled]{algorithm2e}
\usepackage{graphicx}
\usepackage{caption}
\usepackage{subcaption}
\usepackage{booktabs}
\usepackage{makecell}
\usepackage{multirow}
\usepackage{wrapfig}
\usepackage{mathtools}

\usepackage[utf8]{inputenc} % allow utf-8 input
\usepackage[T1]{fontenc}    % use 8-bit T1 fonts
\usepackage{hyperref}       % hyperlinks
\usepackage{url}            % simple URL typesetting
\usepackage{booktabs}       % professional-quality tables
\usepackage{amsfonts}       % blackboard math symbols
\usepackage{nicefrac}       % compact symbols for 1/2, etc.
\usepackage{microtype}      % microtypography
\usepackage{xcolor}         % colors

\title{Relational Object-Centric Actor-Critic}

\author{%
  Leonid Ugadiarov \\
  AIRI \& MIPT, Moscow, Russia \\
  \texttt{ugadiarov@airi.net} \\
  \And
  Vitaliy Vorobyov \\
  FRC CSC RAS \& MIPT, Moscow, Russia \\
  \texttt{vorobev.vitaly.v@phystech.edu} \\
  \And
  Aleksandr Panov \\
  AIRI \& MIPT \& FRC CSC RAS, Moscow, Russia \\
  \texttt{panov@airi.net} \\
}

\begin{document}
\bibliographystyle{plainnat}

\maketitle

\begin{abstract}
The advances in unsupervised object-centric representation learning have significantly improved its application to downstream tasks.
Recent works highlight that disentangled object representations can aid policy learning in image-based, object-centric reinforcement learning tasks. This paper proposes a novel object-centric reinforcement learning algorithm that integrates actor-critic and model-based approaches by incorporating an object-centric world model within the critic. The world model captures the environment's data-generating process by predicting the next state and reward given the current state-action pair, where actions are interventions in the environment.
In model-based reinforcement learning, world model learning can be interpreted as a causal induction problem, where the agent must learn the causal relationships underlying the environment's dynamics. We evaluate our method in a simulated 3D robotic environment and a 2D environment with compositional structure. As baselines, we compare against object-centric, model-free actor-critic algorithms and a state-of-the-art monolithic model-based algorithm. While the baselines show comparable performance in easier tasks, our approach outperforms them in more challenging scenarios with a large number of objects or more complex dynamics.
\end{abstract}

\section{Introduction}

 \begin{figure*}[ht]
   \centering
   \includegraphics[width=0.95\textwidth]{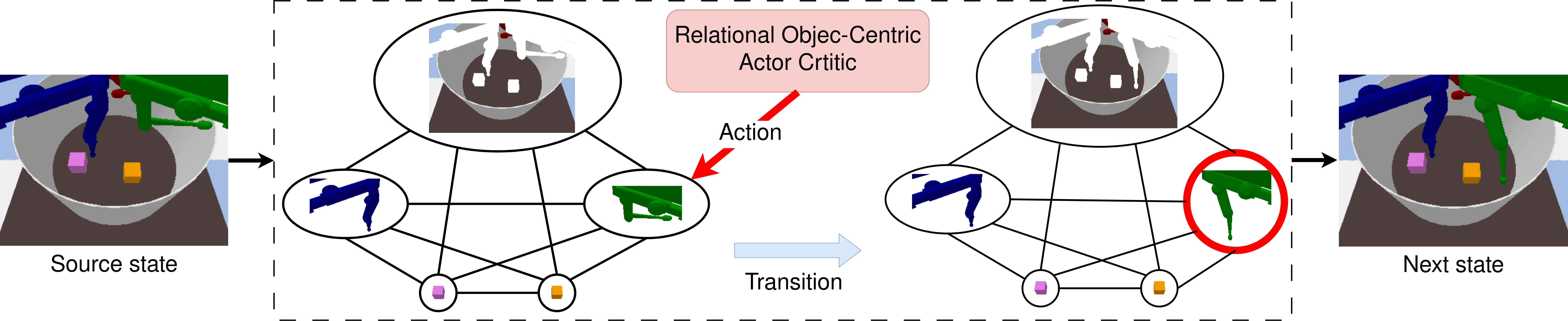}
   \caption{A high-level overview of the proposed method. ROCA learns the policy by extracting object-centric representations from the source image and treating them as a complete graph.} \label{fig:abstract}
 \end{figure*}

One of the primary problems in visual-based reinforcement learning (RL) is determining how to represent the environment's state efficiently. The most common approach is to encode the entire input image, which is then used as input for the policy network~\cite{Mnih2015,zhang2021learning}. However, previous studies~\citep{santoro2017simple} have shown that such representations may fail to capture meaningful relationships and interactions between objects in the state. Object-centric representations (OCR) can be introduced to overcome this issue. Such representations are expected to result in more compact models with enhanced generalization capabilities~\citep{keramati2018fast}. State-of-the-art unsupervised OCR models~\citep{kirilenko_object-centric_2024,singh2022illiterate,kirilenko_quantized_2023,engelcke2022genesisv2} have a fundamental appeal for RL as they do not require additional data labeling for training. Recent studies~\citep{stanić2022learning,yoon2023investigation} have shown that object-centric state factorization can improve model-free algorithms' generalization ability and sample efficiency.

Model-based methods~\citep{10.5555/3312046} represent a promising approach for enhancing data efficiency in RL. In model-based reinforcement learning (MBRL), the agent constructs models of transition and reward functions based on its interactions with the environment. The world model, in this context, describes the underlying data-generating process of the environment by capturing the causal relationships between actions, states, and rewards. By considering actions as interventions in the environment, the learning of the world model can be framed as a causal induction problem~\citep{9363924}. The agent performs multi-step planning to select the optimal action using the model's predictions without interacting with the environment. Model-based algorithms can be more efficient than model-free algorithms, provided the world model's accuracy is sufficiently high. State-of-the-art MBRL methods, which incorporate learning through imagination~\citep{hafner2023mastering} and look-ahead search with a value equivalent dynamics model~\citep{ye2021mastering}, master a diverse range of environments.

To further enhance sample efficiency, a promising direction is to combine both approaches by developing a world model that leverages object representations and explicitly learns to model relationships between objects~\citep{Zholus2022a}. An example of this approach is the contrastively trained transition model CSWM~\citep{Kipf2020Contrastive}. It uses a graph neural network to approximate the dynamics of the environment and simultaneously learns to factorize the state and predict changes in the state of individual objects. CSWM has shown superior prediction quality compared to traditional monolithic models.

However, OCR models demonstrate high quality in relatively simple environments with strongly distinguishable objects~\citep{wu2023slotformer}. Additionally, in object-structured environments, actions are often applied to a single object or a small number of objects, simplifying the prediction of individual object dynamics. In more complex environments, the world model must accurately bind actions to objects to predict transitions effectively. Despite recent progress~\citep{biza2022binding}, no fully-featured dynamics models considering the sparsity of action-object relationships have been proposed. These challenges make it difficult to employ object-centric world models in RL. For instance, the CSWM model has not been utilized for policy learning in offline or online settings.

Our research is focused on value-based MBRL as object-based decomposition of value function could contribute to the training of object-centric world model consistent with policy. From the perspective of extracting causal relationships while learning a world model, an object-centric representation leads to the decomposition of these causal relationships that are responsible for the world model's dynamic part. The model-based approach identifies the effects of agent actions on individual objects and predicts changes in their features or those of small groups. This decomposition simplifies the identified causal relationships, accelerating the learning process for the world model and the agent's policy.

We introduce the Relational Object-Centric Actor-Critic (ROCA), an off-policy object-centric model-based algorithm inspired by the Soft Actor-Critic (SAC) \citep{haarnoja2018soft, haarnoja2019soft, christodoulou2019soft} that operates with both discrete and continuous action spaces. The ROCA algorithm uses the pre-trained SLATE model \citep{singh2022illiterate}, which extracts representations of the individual objects from the input image. Like CSWM \citep{Kipf2020Contrastive}, we utilize a structured transition model based on graph neural networks. Our reward, state-value, and actor models are graph neural networks designed to align with the object-centric structure of the task. Inspired by TreeQN~\citep{farquhar2018treeqn}, we use a world model in the critic module to predict action values. The ROCA algorithm is the first to successfully apply a GNN-based object-centric world model for policy learning in the MBRL setting.

We conducted experiments in 2D environments with simple-shaped objects and visually more complex simulated 3D robotic environments to evaluate the algorithm's quality. The proposed algorithm demonstrates high sample efficiency and outperforms the object-oriented variant of the model-free PPO algorithm~\citep{https://doi.org/10.48550/arxiv.1707.06347}, which uses the same SLATE model as a feature extractor and is built upon the transformer architecture. Furthermore, our method performs better than the state-of-the-art MBRL algorithm DreamerV3~\citep{hafner2023mastering}.

 Our contributions can be summarized as follows:
 \begin{itemize}
     \item We propose a novel architecture that combines a value-based model-based approach with the actor-critic SAC algorithm by incorporating a world model into the critic module.
     \item We extended the SAC algorithm by introducing a new objective function to train the model-based critic.
     \item We propose a GNN-based actor to pool object-centric representations.
     \item We modified the GNN-based CSWM transition model by adjusting its edge model: we pass a pair of slots along with an action into the edge model.
 \end{itemize}

 \begin{figure*}[ht]
   \centering
   \includegraphics[width=0.80\textwidth]{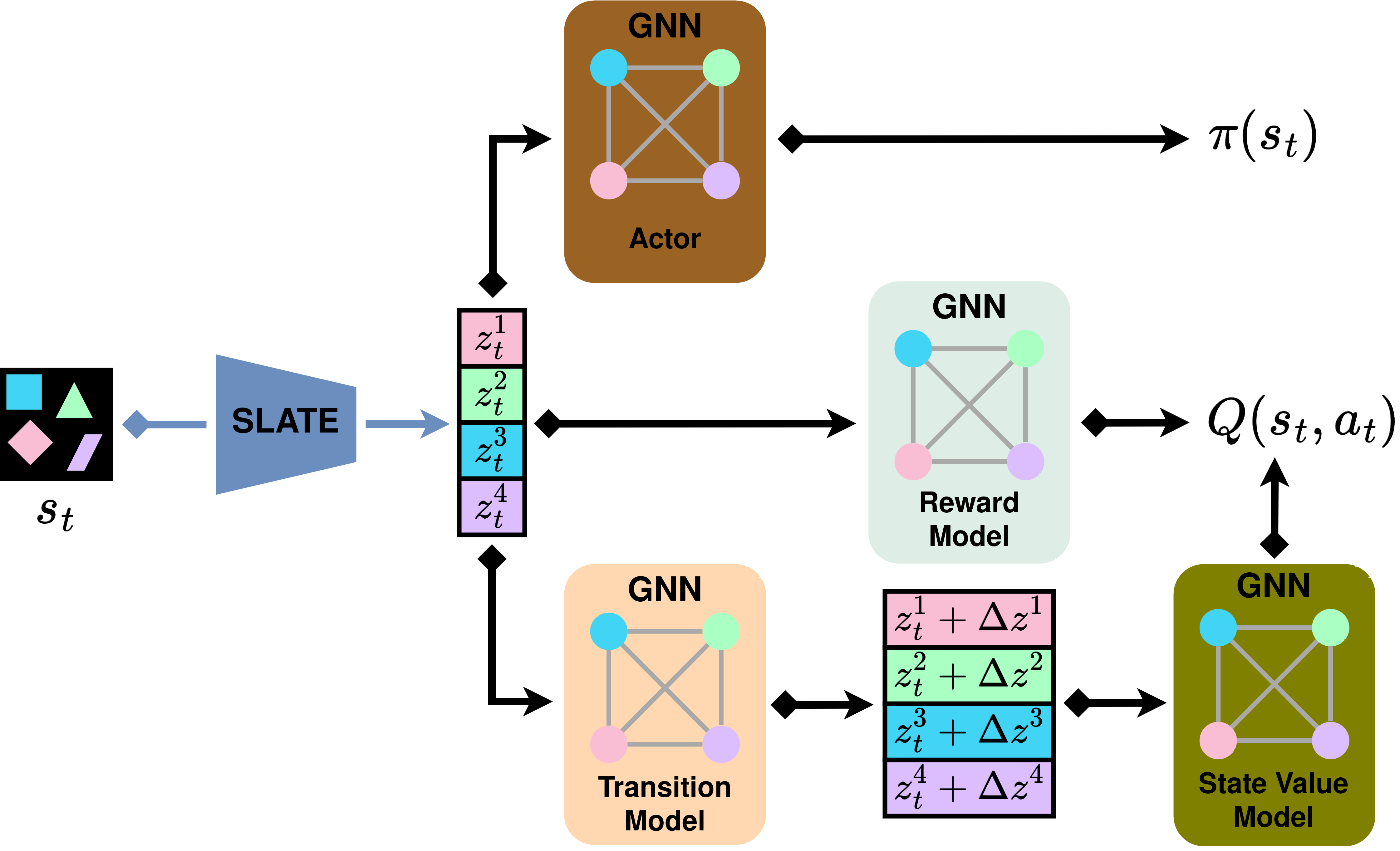}
   \caption{ROCA overview. Framework consists of a pre-trained frozen SLATE model, which extracts object-centric representations from an image-based observation, and GNN-based modules: a transition model, a reward model, a state-value model, and an actor model. The transition and reward models form a world model. The world model and the state-value model together constitute the critic module, which predicts Q-values.} \label{fig:ooqn}
 \end{figure*}

\section{Related Work}
 \paragraph{Object-Centric Representation Learning}
 Recent advancements in machine learning research have been dedicated to developing unsupervised OCR algorithms~\citep{dalle,locatello2020objectcentric,engelcke2022genesisv2}.
 These methods aim to learn structured visual representations from images without relying on labeled data, modeling each image as a composition of objects.
 This line of research is motivated by its potential benefits for various downstream tasks, including enhanced generalization and the ability to reason over visual objects.
 The SPACE~\citep{lin2020spaceunsupervisedobjectorientedscene} method integrates both the spatial attention model and the spatial mixture model. The spatial attention model identifies objects within the scene, whereas the spatial mixture model focuses on separating the remaining background. This combination allows SPACE to effectively differentiate foreground objects and intricate backgrounds.
 One notable approach in this field is Slot-Attention~\citep{locatello2020objectcentric}, which represents objects using multiple latent variables and refines them through an attention mechanism.
 Building upon this, SLATE~\citep{dalle} further improves the performance by employing a Transformer-based decoder instead of a pixel-mixture decoder.
 Another recently proposed approach that doesn't rely on Slot-Attention is the Deep Latent Particles (DLP) method~\citep{daniel2024ddlpunsupervisedobjectcentricvideo}.
 DLP encodes an input image into a disentangled latent space, structured as a set of particles with interpretable attributes.
 These include the particle's position, scale, a "depth" attribute in pixel space to determine which particle is in front in case of overlap, a transparency attribute, and latent features that capture the visual appearance of the region surrounding each particle.
 
 \paragraph{Object-Centric Representations and Model-Free RL}
 \citet{stanić2022learning} uses Slot-Attention as an object-centric feature extractor and examines the performance and generalization capabilities of RL agents.
 In another study~\citep{sharma2023objectcentric}, a multistage training approach is proposed, involving fine-tuning a YOLO model~\citep{https://doi.org/10.5281/zenodo.3908559} on a dataset labeled by an unsupervised object-centric model.
 The frozen YOLO model is then employed as an object-centric features extractor in the Dueling DQN algorithm.
 Object representations are pooled using a graph attention neural network before being fed to the Q-network.
\citet{yoon2023investigation} proposes using a transformer encoder as a pooling layer in the PPO ~\citep{https://doi.org/10.48550/arxiv.1707.06347} to aggregate the object representations extracted from an input image by object-centric models of different types.
\citet{yi2022objectcategoryawarereinforcementlearning} uses a pre-trained SPACE model to extract object-centric representations and unsupervisedly groups them into a set of categories.
For each category, an independent neural network module is used in the PPO, improving the sample efficiency of the approach compared to both monolithic and object-centric baselines.
SMORL~\citep{zadaianchuk2020selfsupervisedvisualreinforcementlearning} and SRICS~\citep{zadaianchuk2022selfsupervisedreinforcementlearningindependently} are object-centric, model-free methods specifically designed for goal-conditioned manipulation tasks. SRICS utilizes GNNs as transition models but is limited to non-visual observations. SMORL is suitable for visually-based tasks, as it uses a patch-based image representation model, SCALOR~\citep{jiang2020scalorgenerativeworldmodels}. However, its key limitation is the assumption that multi-object tasks can be addressed sequentially and independently, ignoring potential interactions between objects.
\citet{haramati2024entitycentricreinforcementlearningobject} combines DLP and TD3~\citep{fujimoto2018addressingfunctionapproximationerror} with a transformer backbone for goal-conditioned manipulation tasks.
Additionally, the authors design an intrinsic reward based on the disentangled features of objects provided by DLP.
The proposed method outperforms state-of-the-art goal-based approaches.

 \paragraph{Object-Centric Representations and MBRL}
 As related work in object-oriented MBRL, we consider \cite{watters2019cobra}.
 It uses MONet~\citep{burgess2019monet} as an object-centric features extractor and learns an object-oriented transition model.
 However, unlike our approach, this model does not consider the interaction between objects and is only utilized during the exploration phase of the RL algorithm.
 Recently proposed MBRL algorithm FOCUS~\citep{ferraro2023focusobjectcentricworldmodels} is based on a modified version of RSSM~\citep{hafner2022masteringataridiscreteworld} and incorporates an online-learning object-centric encoder.
 FOCUS accelerates training by using an exploration bonus and requires segmentation masks for its decoder.
 These masks are either provided by the simulator in a supervised regime or generated by the Segment Anything Model~\citep{kirillov2023segment} after fine-tuning with ground-truth segmentation masks in a weakly supervised regime.

\section{Background}
 \subsection{Markov Decision Process}
 We consider a simplified version of the object-oriented MDP \citep{10.1145/1390156.1390187}:
 \begin{equation} \label{eq:MDP}
     \mathcal{U} = (\mathcal{S}, \mathcal{A}, T, R, \gamma, \mathcal{O}, \Omega),
 \end{equation}
 where $\mathcal{S} = \mathcal{S}_1 \times \dots \times \mathcal{S}_K$ --- a state space, $\mathcal{S}_i$ --- an individual state space of the object $i$, $\mathcal{A}$ --- an action space, $T = (T_1, \dots, T_K)$ --- a transition function, $T_i = T_i(T_{i 1}(s_i, s_1, a), \dots, T_{i K}(s_i, s_K, a))$ --- an individual transition function of the object $i$, $R = \sum_{i = 1} ^ K R_i$ --- a reward function, $R_i = R_i(R_{i 1}(s_i, s_1, a), \dots, R_{i K}(s_i, s_K, a))$ --- an individual reward function of the object $i$, $\gamma \in \left[ 0; 1 \right]$ --- a discount factor, $\mathcal{O}$ --- an observation space, $\Omega: \mathcal{S} \rightarrow \mathcal{O}$ --- an observation function.
 The goal of reinforcement learning is to find the optimal policy: \\ $\pi ^ * = \text{argmax}_{\pi} \mathbb{E}_{s_{t + 1} \sim T(\cdot \vert s_t, a_t), a_{t + 1} \sim \pi(\cdot \vert s_{t + 1}) } \left[ \sum_{i = 0} ^{\tau} \gamma ^ {t} R(s_t, a_t) \right]$ for all $s_0$ where $\tau$ is the number of time steps.

 In model-based approach the agent uses the experience of interactions with the environment to build a world model that approximates the transition function $\hat{T} \approx T$ and the reward function $\hat{R} \approx R$ and use its predictions as an additional signal for policy learning.

 \subsection{Soft Actor-Critic}
 Soft Actor-Critic (SAC) \citep{haarnoja2018soft, haarnoja2019soft} is a state-of-the-art off-policy reinforcement learning algorithm for continuous action settings.
 The goal of the algorithm is to find a policy that maximizes the maximum entropy objective:
 \begin{equation*}
 \begin{split}
          \pi ^ * = \text{argmax}_{\pi} \sum_{i = 0} ^ {\tau} \mathbb{E}_{(s_t, a_t) \sim d_{\pi} } \big[  \gamma ^ {t} (R(s_t, a_t) + \alpha \mathcal{H}(\pi(\cdot|s_t))\big]
 \end{split}
 \end{equation*}
 where $\alpha$ is the temperature parameter, $\mathcal{H}(\pi(\cdot|s_t)) = -\log{\pi(\cdot|s_t)}$ is the entropy of the policy $\pi$ at state $s_t$, $d_{\pi}$ is the distribution of trajectories induced by policy $\pi$.
 The soft action-value function $Q_{\theta}(s_t, a_t)$ parameterized using a neural network with parameters $\theta$ is trained by minimizing the soft Bellman residual:
 \begin{equation}\label{eq:sac_critic_objective}
 \begin{split}
 J_{Q}(\theta) = \mathbb{E}_{(s_t, a_t) \sim D} \big[ \big( Q_{\theta}(s_t, a_t) - R(s_t, a_t) - \gamma \mathbb{E}_{s_{t + 1} \sim T(s_t, a_t)} V_{\bar{\theta}}(s_{t + 1}) \big) ^ 2 \big]
 \end{split}
 \end{equation}
 where $D$ is a replay buffer of past experience and $V_{\bar{\theta}}(s_{t + 1})$ is estimated using a target network for $Q$ and a Monte Carlo estimate of the soft
 state-value function after sampling experiences from the $D$.

 The policy $\pi$ is parameterized using a neural network with parameters $\phi$. The parameters are learned by minimizing the expected KL-divergence between the policy and the exponential of the $Q$-function:
 \begin{equation}\label{eq:sac_actor_objective}
 \begin{split}
     J_{\pi}(\phi) = \mathbb{E}_{s_t \sim D} \big[ \mathbb{E}_{a_t \sim \pi_{\phi}(\cdot|s_t)} \big[\alpha \log(\pi_{\phi}(a_t|s_t)) - Q_{\theta}(s_t, a_t) \big]\big]
 \end{split}
 \end{equation}

 The objective for the temperature parameter is given by:
 \begin{equation}\label{eq:sac_temperature_continuous}
     J(\alpha) = \mathbb{E}_{a_t \sim \pi(\cdot|s_t)} \big[ -\alpha (\log \pi(a_t|s_t) + \bar{H}) \big]
 \end{equation}
 where $\bar{H}$ is a hyperparameter representing the target entropy.
 In practice, two separately trained soft Q-networks are maintained, and then the minimum of their two
 outputs are used to be the soft Q-network output.

 While the original version of SAC solves problems with continuous action space, the version for discrete action spaces was suggested by \citet{christodoulou2019soft}.
 In the case of discrete action space, $\pi_{\phi}(a_t|s_t)$ outputs a probability for all actions instead of a density.
 Such parametrization of the policy slightly changes the objectives \ref{eq:sac_critic_objective}, \ref{eq:sac_actor_objective} and \ref{eq:sac_temperature_continuous}.

\section{Relational Object-Centric Actor-Critic}
 Figure \ref{fig:ooqn} outlines the high-level overview of the proposed actor-critic framework (ROCA).
 As an encoder we use SLATE~\citep{singh2022illiterate}, a recent object-centric model. SLATE incorporates a dVAE \citep{oord2018neural} for internal feature extraction, a GPT-like transformer \citep{dalle} for decoding, and a slot-attention module \citep{slotattention} to group features associated with the same object.
 We refer to the appendix~\ref{appendix:slate} for a more detailed description of SLATE.
 In ROCA the pre-trained frozen SLATE model takes an image-based observation $s_t$ as input and produces a set of object vectors, referred to as slots, $\boldsymbol{z}_t = (z_t ^ 1, \dots, z_t ^ K)$ ($K$ - the maximum number of objects to be extracted).
 An actor model encapsulates the current agent's policy and returns an action for the input state $\boldsymbol{z}_t$. 
 Critic predicts the value $Q(\boldsymbol{z}_t, a)$ of the provided action $a$ sampled from the actor given the current state representations $\boldsymbol{z}_t$.
 It is estimated using the learned transition model, reward model, and state-value model.
 The input state representation $\boldsymbol{z}_t = (z_t ^ 1, \dots, z_t ^ K)$ is treated as a complete graph while being processed by GNN-based components of the ROCA.

 \subsection{Transition Model}
 We approximate the transition function using a graph neural network \cite{Kipf2020Contrastive} with an edge model $\texttt{edge}_T$ and a node model $\texttt{node}_T$ which takes a factored state $\boldsymbol{z}_t = (z_t ^ 1, \dots, z_t ^ K)$ and action $\boldsymbol{a}_t$ as input and predicts changes in factored states $\Delta \boldsymbol{z}$.
 The action is provided to the node model $\texttt{node}_T$ and the edge model $\texttt{edge}_T$ as shown in the appendix in Figure \ref{fig:gnn}.
 The factored representation of the next state is obtained via $\hat{\boldsymbol{z}}_{t + 1} = \boldsymbol{z}_t + \Delta \boldsymbol{z}$.
 Since we treat the set of slots as a complete graph, the complexity of the update rule (\ref{eq:transition}) is quadratic in the number of slots.
 The same applies to all GNN models in the ROCA.

 \begin{equation} \label{eq:transition}
     \Delta z ^ i = \texttt{node}_{T} ( {z}_t ^ i, {a}_t ^ i, \sum_{i \neq j} \texttt{edge}_T (z_t ^ i, z_t ^ j, {a_t ^ i}))
 \end{equation}

 \subsection{Reward Model}
 The reward model uses almost the same architecture as the transition model. Still, we average object embeddings returned by the node models and feed the result into the MLP to produce the scalar reward.
 The reward model is trained using the mean squared error loss function with environmental rewards $r_t$ as target (\ref{eq:reward}).
 \begin{equation} \label{eq:reward}
     \begin{cases}
     \texttt{embed}_R ^ i \mkern-18mu &= \texttt{node}_{R} ( {z}_t ^ i, {a}_t ^ i, \sum_{i \neq j} \texttt{edge}_R (z_t ^ i, z_t ^ j, {a_t ^ i})) \\
     \hat{R}(\boldsymbol{z}_t, {a}_t) \mkern-18mu &= MLP(\sum_{i = 1} ^ K \texttt{embed}_R ^ i / K) \\
     \end{cases}
 \end{equation}

 \subsection{State-Value Model}
 The state-value function is approximated using a graph neural network $\hat{V}$, which does not depend on actions in either the edge model $\texttt{edge}_V$ or the node model $\texttt{node}_V$.
 As in the reward model, we average object embeddings returned by the node models and feed the result into the MLP to produce the scalar value.
 \begin{equation} \label{eq:value}
     \begin{cases}
     \texttt{embed}_V ^ i \mkern-18mu &= \texttt{node}_{V} ( {z}_t ^ i, \sum_{i \neq j} \texttt{edge}_V (z_t ^ i, z_t ^ j)) \\
     \hat{V}(\boldsymbol{z}_t) \mkern-18mu &= MLP(\sum_{i = 1} ^ K \texttt{embed}_V ^ i / K) \\ 
     \end{cases}
 \end{equation}

 \subsection{Actor Model}
 The actor model uses the same GNN architecture as the state-value model but employs different MLP heads for continuous and discrete action spaces. In the case of the continuous action space, it returns the mean and the covariance of the Gaussian distribution. For the discrete action space, it outputs the probabilities for all actions.
 \begin{equation} \label{eq:actor}
     \begin{cases}
     \texttt{embed}_{actor} ^ i \mkern-18mu &= \texttt{node}_{actor} ( {z}_t ^ i, \sum_{i \neq j} \texttt{edge}_{actor} (z_t ^ i, z_t ^ j)) \\
     \mu(\boldsymbol{z}_t) \mkern-18mu &= MLP_{\mu}(\sum_{i = 1} ^ K \texttt{embed}_{actor} ^ i / K) \\
     \sigma ^ 2 (\boldsymbol{z}_t) \mkern-18mu &= MLP_{\sigma ^ 2}(\sum_{i = 1} ^ K \texttt{embed}_{actor} ^ i / K) \\
     \pi (\boldsymbol{z}_t) \mkern-18mu &= MLP_{\pi}(\sum_{i = 1} ^ K \texttt{embed}_{actor} ^ i / K) \\
     \end{cases}
 \end{equation}

 \subsection{Critic Model}
 In the critic, we use a world model to predict action-values.
 Specifically, we employ a Q-function decomposition based on the Bellman equation. It was initially introduced in the Q-learning TreeQN algorithm~\citep{farquhar2018treeqn}:
 \begin{equation} \label{eq:critic}
     \hat{Q} (\boldsymbol{z}_{t}, {a}_{t}) = \hat{R} (\boldsymbol{z}_{t}, {a}_{t}) + \gamma \hat{V} (\boldsymbol{z}_t + \Delta \boldsymbol{z})
 \end{equation}
 where $\hat{R}$ --- the reward model (\ref{eq:reward}), $\hat{V}$ --- the state-value model (\ref{eq:value}), $\boldsymbol{z}_t + \Delta \boldsymbol{z}$ --- the next state prediction, generated by the transition model (\ref{eq:transition}).
 Since the critic's output values are computed using the world model, we refer to our approach as a value-based model-based method.

 \begin{figure*}[ht]
   \centering
   \includegraphics[width=\textwidth]{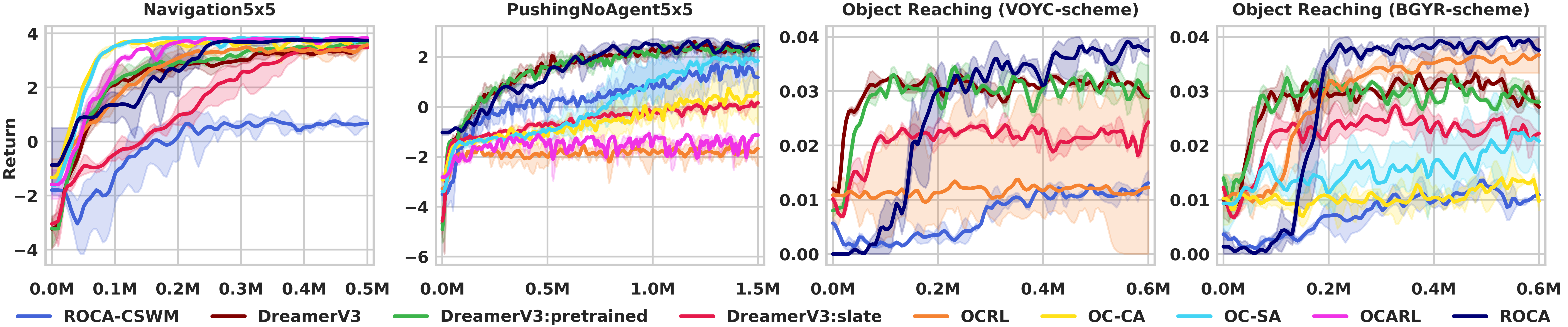}
   \caption{Return averaged over 30 episodes and three seeds for ROCA, ROCA-CSWM, DreamerV3, DreamerV3:pretrained, DreamerV3:slate, OCRL, OC-CA, OC-SA, and OCARL. ROCA learns faster or achieves higher metrics than the baselines. Shaded areas indicate standard deviation.} \label{fig:results}
 \end{figure*}

 \subsection{Training}
 The SLATE model is pre-trained on the data set of trajectories collected with a uniform random policy (300K observations for Shapes2D tasks and 1M observations for the Object Reaching task). Following the original paper \citep{singh2022illiterate}, we apply decay on the dVAE temperature $\tau$ from 1.0 to 0.1 and a learning rate warm-up for the parameters of the slot-attention encoder and the transformer at the start of the training.
 After pre-training, we keep the parameters of the SLATE model frozen.

 To train all the other components of ROCA we use SAC objectives (\ref{eq:sac_critic_objective}, \ref{eq:sac_actor_objective}, \ref{eq:sac_temperature_continuous}).
 For both continuous and discrete environments, a conventional double Q-network architecture is used in the critic module. Additionally, we use the data sampled from the replay buffer to train the world model components. The transition model is trained using the mean squared error loss function to minimize the prediction error of the object representations for the next state, given the action. The reward model is trained using the mean squared error loss function with environmental rewards $r_t$ as targets.
 \begin{equation}\label{eq:wm_objective}
 \begin{split}
     J_{WM} = \mathbb{E}_{s_t, a_t, r_t, s_{t + 1} \sim D} \big[ \beta_{T} \lVert \boldsymbol{z}_t + \Delta \boldsymbol{z} - \boldsymbol{z}_{t + 1} \rVert ^ 2 + \beta_{R} \big(\hat{R} (\boldsymbol{z}_t, {a}_t) - r_t \big) ^ 2 \big]
 \end{split}
 \end{equation}
 In total, we use four optimizers. The temperature parameter, the actor, and the value model use individual optimizers. The transition and reward models share the world model optimizer.

 Due to the stochastic nature of the SLATE model, object-centric representation can shuffle at each step. To enforce the order of object representation during the world model objective (\ref{eq:wm_objective}) optimization, we pre-initialize the slots of the SLATE model for the next state $\boldsymbol{z}_{t+1}$ with the current values $\boldsymbol{z}_{t}$.

\section{Environments}
 The efficiency of the proposed ROCA algorithm was evaluated in the 3D robotic simulation environment CausalWorld \citep{ahmed2020causalworld} on the Object Reaching task as it was done in \citep{yoon2023investigation}, and in the compositional 2D environment Shapes2D \citep{Kipf2020Contrastive} on the Navigation and PushingNoAgent tasks.
 Current state-of-the-art slot-based object-centric models struggle to extract meaningful object-centric representations in visually complex environments~\citep{NEURIPS2020_8511df98, NEURIPS2021_43ec517d}.
 As a result, testing object-centric RL algorithms in visually rich environments, like Habitat~\citep{szot2022habitat}, becomes challenging due to the low quality of representations.
 However, the visual complexity of the selected environments enables object-centric models to extract high-quality object representations.
 This allows us to focus on the problem of object-centric MBRL, which is the primary objective of this paper.

 \paragraph{Object Reaching Task} 
 In this task, a fixed target object (violet cube) and a set of distractor objects (orange, yellow, and cyan cubes) are randomly placed in the scene.
 The agent controls a tri-finger robot and must reach the target object with one of its fingers (the other two are permanently fixed) to obtain a positive reward and solve the task.
 The episode ends without reward if the finger first touches one of the distractor objects.
 The action space in this environment consists of the three continuous joint positions of the moveable finger.
 During our experiments, we discovered that one of the baseline algorithms is sensitive to the choice of color scheme for the cubes.
 Therefore, we also conducted experiments in the task with the original color scheme~\citep{yoon2023investigation}: the color of the target cube is blue, and the colors of the distracting cubes are red, yellow, and green.

 \paragraph{Navigation Task}
 Shapes2D environment is a four-connected grid world where objects are represented as figures of simple shapes.
 One object --- the cross is selected as a stationary target. The other objects are movable.
 The agent controls all movable objects.
 In one step, the agent can move an object to any free adjacent cell.
 The agent aims to collide the controlled objects with the target object.
 Upon collision, the object disappears, and the agent receives a reward of $+1$.
 When an object collides with another movable object or field boundaries, the agent receives a reward of $-0.1$, and the positions of objects are not changed.
 For each step in the environment, the agent receives a reward of $-0.01$.
 The episode ends if only the target object remains on the field.
 In the experiments, we use a 5x5-sized environment with five objects and a 10x10-sized environment with eight objects. The action space in the Shapes2D environment is discrete and consists of 16 actions for the Navigation 5x5 task (four movable objects) and 28 actions for the Navigation 10x10 task (seven movable objects).
 \paragraph{PushingNoAgent Task} The agent controls all movable objects as in the Navigation task, but collisions between two movable objects are permitted: both objects move in the direction of motion.
 The agent is tasked to push another movable object into the target while controlling the current object.
 The pushed object disappears, and the agent receives a reward of $+1$ for such an action.
 When the currently controlled object collides with the target object or field boundaries, the agent receives a reward of $-0.1$.
 When the agent pushes a movable object into the field boundaries, the agent receives a reward of $-0.1$.
 For each step in the environment, the agent receives a reward of $-0.01$.
 The episode ends if only the target object and one movable object remain on the field.
 In the experiments, we use a 5x5-sized environment with five objects.

 \begin{figure}[ht]
   \centering
     \includegraphics[width=0.47\linewidth]{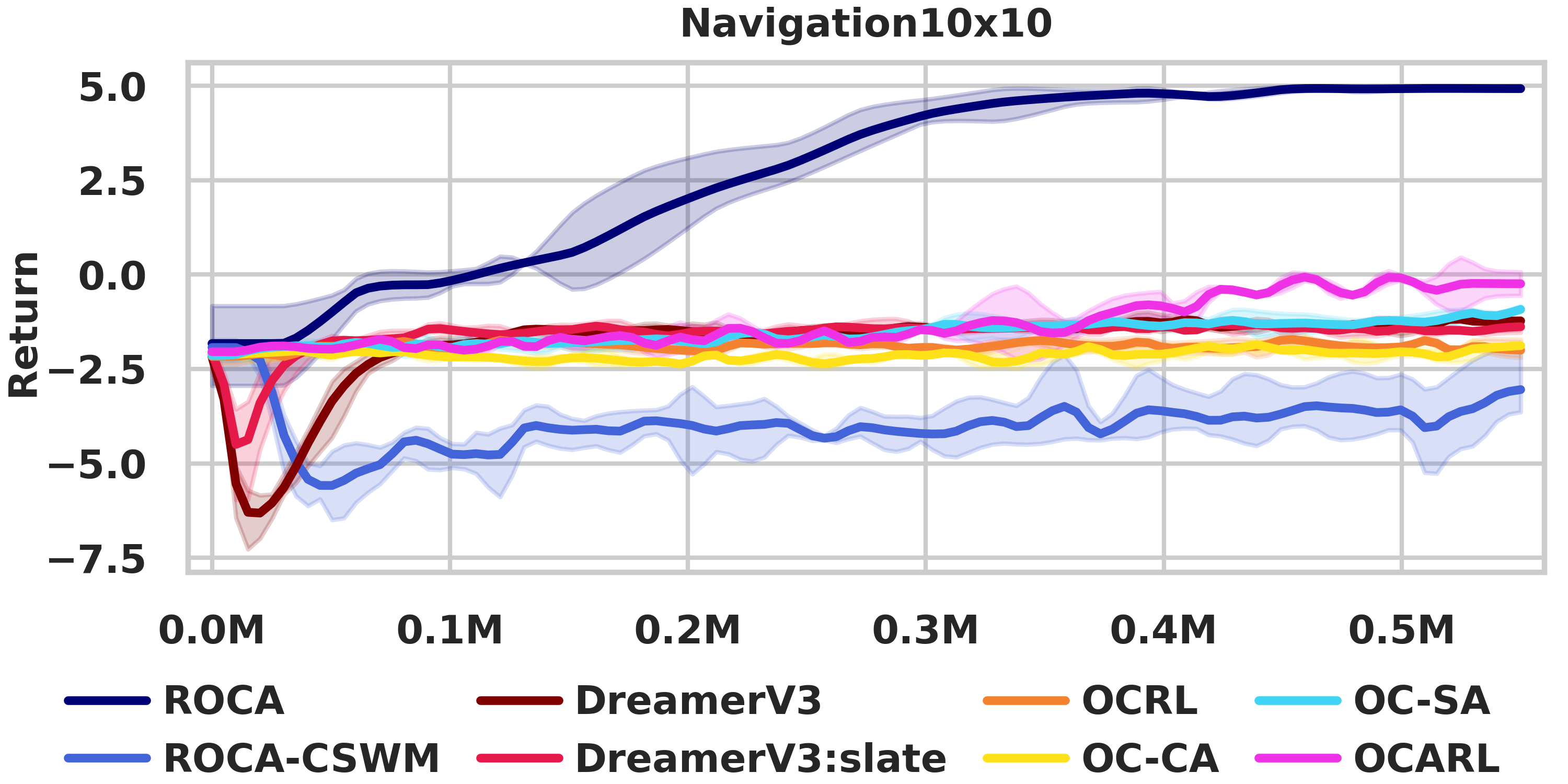}
     \caption{Return averaged over 30 episodes and three seeds for ROCA, ROCA-CSWM, DreamerV3, DreamerV3:slate, OCRL, OC-SA, OC-CA, and OCARL models in the Navigation 10x10 task. ROCA exhibits better performance than baselines but still does not solve the task. Shaded areas indicate standard deviation.}
 % \hspace{-3cm}
 \label{fig:results_nav10x10}
 \end{figure}

\section{Experiments}
 We utilize a single SLATE model for Navigation5x5 and PushingNoAgent5x5 tasks as they share the same observation space.
 However, we train a distinct SLATE model for Navigation10x10 and each version of the Object Reaching task. 
 The appendix provides detailed information regarding the hyperparameters of the SLATE model in \ref{appendix:slate} and the examples of attention maps produced by SLATE in Figure \ref{fig:appendix_slots}.

 In continuous Object Reaching tasks, we conventionally use the dimension of the action space as the target entropy hyperparameter for ROCA.
 For 2D tasks with a discrete action space, we scale the entropy of a uniform random policy with the tuned coefficient.
 For more information on the hyperparameters of the ROCA model, please refer to the appendix \ref{appendix:roca}.

 We compare ROCA with an object-centric model-free algorithm based on PPO, using the same pre-trained frozen SLATE model as a feature extractor.
 To combine the latent object representations into a single vector suitable for the value and policy networks of the PPO, we used a Transformer encoder~\citep{vaswani2023attention} as a pooling layer.
 We referred to the transformer-based PPO implementation provided by \citep{yoon2023investigation} as the OCRL baseline.
 For the Object Reaching Task, we employed the same hyperparameter values as the authors.
 For Shapes2D tasks, we fine-tuned the hyperparameters of the OCRL baseline.
The other object-centric model-free baselines are the self-attention OC-SA and cross-attention OC-CA versions from \citet{stanić2022learning}, as well as OCARL~\citep{yi2022objectcategoryawarereinforcementlearning}.
The tested hyperparameters are listed in the appendix.
 
 Since there are no established state-of-the-art object-centric MBRL algorithms, we have chosen the DreamerV3~\citep{hafner2023mastering} algorithm as a MBRL baseline.
 In order to ensure a fair comparison between the ROCA and the DreamerV3, we conducted experiments where we trained the DreamerV3 with a pretrained encoder obtained from the DreamerV3 model that solves the task (DreamerV3:pretrained).
 For all the tasks, we conducted experiments using two different modes: one with the encoder frozen and another with the encoder unfrozen.
 However, we did not observe any improvement in the convergence rate compared to the DreamerV3 model that does not use the pretrained encoder.
 Additionally, we discovered that the pretrained world model significantly accelerates the convergence of DreamerV3, but this mode makes the comparison unfair to the ROCA.
 We also utilize the pretrained frozen SLATE model as an encoder in DreamerV3.
 In DreamerV3:slate, we obtain a single-vector representation of the current observation by concatenating the slot representations produced by SLATE and then feed it into DreamerV3.
 For the DreamerV3-based algorithms we use the default hyperparameter values from the official repository, except for the \texttt{train\_ratio} parameter which we increased from 32 to 512, as we observed that more frequent training of models improves sample efficiency.
 Additionally, we implement ROCA-CSWM baseline, where we utilize CNN-based object encoder and train it online along with the world model using contrastive loss~\cite{Kipf2020Contrastive}: 
 \begin{equation}\label{eq:wm_contrastive}
 \begin{split}
     J_{WM} ^ {cswm} &= \mathbb{E}_{s_t, a_t, r_t, s_{t + 1} \sim D} \big[ \beta_{T} \lVert \boldsymbol{z}_t + \Delta \boldsymbol{z} - \boldsymbol{z}_{t + 1} \rVert ^ 2 \\
     & \quad + \beta_{R} \big(\hat{R} (\boldsymbol{z}_t, {a}_t) - r_t \big) ^ 2 + \beta_{C} \text{max} \big( 0, \gamma - \lVert \hat{\boldsymbol{z}}_t -  \boldsymbol{z}_{t + 1} \rVert ^ 2\big)
     \big],
 \end{split}
 \end{equation}
 where $\gamma$ is the margin and $\hat{\boldsymbol{z}}_t$ is the representation of the observation randomly sample from the replay buffer. Other loss functions in ROCA and ROCA-CSWM are the same.

 The hyperparameters for ROCA-CSWM can be found in the appendix \ref{appendix:roca-cswm}.

 \paragraph{Results}

 \begin{figure*}[ht]
   \centering
   \includegraphics[width=\textwidth]{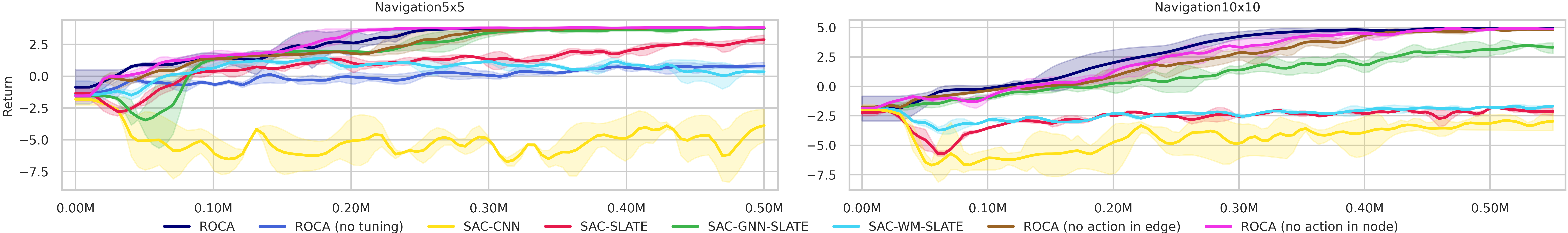}
   \caption{Ablation study. SAC-CNN --- a version of SAC with a standard CNN encoder. SAC-SLATE --- a version of SAC with a pretrained SLATE encoder which averages object emebeddings to obtain the embedding of the current state. SAC-WM-SLATE --- a modification of SAC-SLATE which uses a monolithic world-model in its critic. SAC-GNN-SLATE --- an object-centric version of SAC with a pretrained SLATE encoder which uses GNNs as actor and critic. ROCA (no-tuning) --- a version of ROCA without target entropy tuning. ROCA (no action in edge) --- a version of ROCA, in which \texttt{edge} functions do not take an action as input. ROCA (no action in node) --- a version of ROCA, in which \texttt{node} functions do not take an action as input. ROCA and ROCA (no action in node) demonstrate similar performance. ROCA outperforms the other considered baselines. Return averaged over 30 episodes and three seeds. Shaded areas indicate standard deviation.} \label{fig:results_ablations}
 \end{figure*}

 \begin{figure*}[ht]
   \centering
     \includegraphics[width=0.65\linewidth]{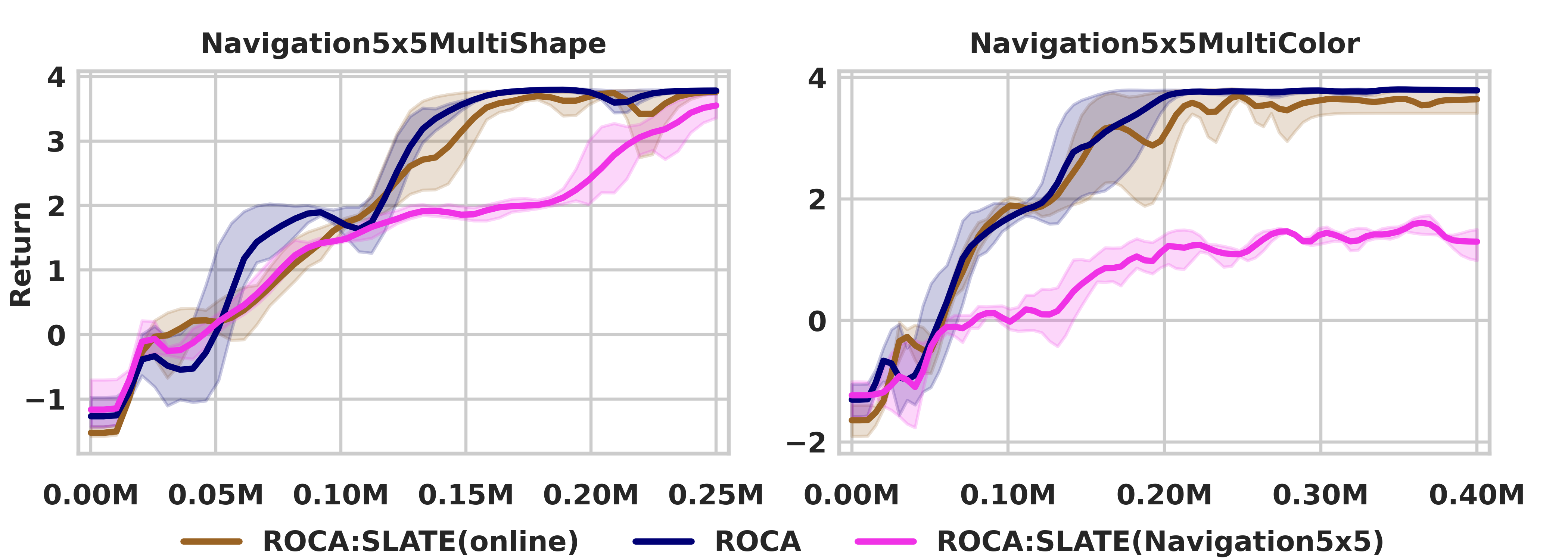}
     \caption{Return averaged over 30 episodes and three seeds for ROCA, ROCA:SLATE(online), ROCA:SLATE(Navigation5x5) in the modified versions of the Navigation5x5 task. These tasks introduce distribution shifts for the frozen SLATE used in ROCA:SLATE(Navigation5x5). ROCA and ROCA:SLATE(online) perform comparably, whereas ROCA:SLATE(Navigation5x5) fails to solve the Navigation5x5MultiColor task. Shaded areas indicate standard deviation.}
 % \hspace{-3cm}
 \label{fig:results_ood}
 \end{figure*}

 The graphs in Figure \ref{fig:results} depict how the episode return of ROCA and the baselines depend on the number of steps for Navigation 5x5, PushingNoAgent5x5, and two versions of the Object Reaching task. 
 For the Navigation 5x5 task, ROCA performs better than the OCRL baseline.
 Although DreamerV3 shows slightly more stable and efficient learning than ROCA, ROCA eventually achieves a higher return.
 In the PushingNoAgent 5x5 task, ROCA and DreamerV3 demonstrate similar performance.
 For the Object Reaching task with the original color scheme (BGYR), the OCRL baseline initially demonstrates much better performance, but ROCA surpasses all baselines after 200K steps.
 We believe that the poor performance of the OCRL baseline in the Object Reaching task with VOYC color schema is due to its sensitivity to the quality of the SLATE model.
 One potential solution to overcome this issue could be increasing the number of training epochs for the SLATE.
 Figure \ref{fig:results_nav10x10} demonstrates the results in the more challenging Navigation 10x10 task. All baselines fail to achieve a positive return.
 ROCA performs better than the baselines but can not solve the task entirely, as it only moves five out of seven objects to the target.
 In all tasks, DreamerV3 outperforms DreamerV3:slate, which indicates that DreamerV3 cannot efficiently utilize object-centric representations.
 The results for ROCA-CSWM highlight the difficulties of using contrastive loss for training the world model and the object encoder in an online setting.

 \paragraph{Ablations}
 ROCA is built upon SAC, and thus, the ablation study aims to assess the impact of the different modifications we introduced to the original SAC with a monolithic CNN encoder.
 Figure \ref{fig:results_ablations} illustrates the results of additional experiments estimating the effects of the pre-trained SLATE encoder, the object-centric actor and critic, the object-centric world model and the target entropy tuning.
 We evaluate the quality of several monolithic and object-centric versions of SAC and compare them with ROCA.
 SAC-CNN is standard monolithic version of SAC that utilizes the convolutional encoder from the original DQN implementation~\citep{Mnih2015}.
 In SAC-SLATE, the CNN encoder is replaced with a pre-trained frozen SLATE encoder, while the other model components remain the same.
 To obtain the monolithic state representation $z_t ^ {*}$ from the object-centric one $\boldsymbol{z}_t$, produced by the SLATE, we take the average over the object axis: $z_t ^ {*} = \sum_{i = 0} ^ {K} z_t ^ i / K$.
 Note, that $z_t ^ {*}$ is independent of the slot order in $\boldsymbol{z}_t$ and can be fed into the standard actor and critic MLPs.
 SAC-WM-SLATE builds upon SAC-SLATE and can be considered as a monolithic version of the ROCA.
 Its actor, state-value, reward, and transition models are implemented using MLPs.
 SAC-GNN-SLATE is an object-centric version of SAC and can be viewed as ROCA without the world model in the critic module.
 It uses a pretrained frozen SLATE encoder and GNN-based actor and critic modules.
 Additionally, we compare the ROCA with a variant where the target entropy is set to the default value, equal to the scaled entropy of the uniform random policy with coefficient 0.98~\citep{christodoulou2019soft}.
 Also we compare ROCA with a variant ROCA (without actions in GNN edges) in which the \texttt{edge} functions of the transition (\ref{eq:transition}) model and the reward model (\ref{eq:reward}) do not take the action $a_t$ as input.
 This architecture strictly follows the CSWM model~\citep{Kipf2020Contrastive}.

 The ablation studies have shown that in the monolithic mode, the SLATE model significantly improves performance only in the relatively simple Navigation5x5 task.
 However, extending the critic with the world model does not improve the convergence rate.
 The object-centric SAC-GNN-SLATE outperforms all monolithic models.
 Finally, the ROCA, which uses an object-centric world model in the critic module, outperforms the SAC-GNN-SLATE.
 Note that we obtained the presented results after fine-tuning the hyperparameters for all of the models.
 Also we observe that providing actions to the \texttt{edge} function in GNNs slightly improve performance of the our algorithm.

\paragraph{Performance with Frozen SLATE Under Distribution Shifts}
We conducted two experiments to investigate the performance of our approach under distribution shifts.
Specifically, we used SLATE trained on the Navigation5x5 task but train ROCA on modified versions of Navigation5x5.
In the Navigation5x5MultiShape task, the shapes of three objects were randomly sampled at the beginning of each episode (with a probability of 0.5 for each object): the red circle could be replaced with a red diamond, the blue triangle with a blue pentagon, and the green square with a green scalene triangle.
In the Navigation5x5MultiColor task, the colors of these objects were randomly sampled at the beginning of each episode (also with a probability of 0.5 for each object): the red circle could be replaced with a yellow circle, the blue triangle with a brown triangle, and the green square with a pink square.
Both Navigation5x5MultiShape and Navigation5x5MultiColor introduce distribution shifts compared to the original Navigation5x5 task.
We evaluated three versions of ROCA in these experiments.
ROCA:SLATE(Navigation5x5) --- ROCA with a frozen SLATE trained on data collected from the original Navigation5x5 task.
ROCA:SLATE(online) --- ROCA with SLATE initialized with weights from the Navigation5x5 task but trained further online along with the policy on data from the replay buffer.
ROCA --- The standard version of ROCA with a frozen SLATE trained on data collected from the current task.

The results, presented in Figure \ref{fig:results_ood}, show that ROCA converges the fastest, as expected, since it does not experience any distribution shift.
ROCA:SLATE(online) learns slightly slower but successfully converges in both tasks.
In contrast, ROCA:SLATE(Navigation5x5) learns the slowest and only solves the Navigation5x5MultiShape task, failing to converge in Navigation5x5MultiColor within 400K steps.
From these results, we conclude that ROCA can adapt to distribution shifts between offline and online data by training SLATE alongside the policy on data from the replay buffer.

The results of additional experiments with DreamerV3 using a pre-trained frozen encoder, sparsification of the GNNs in ROCA, evaluation of ROCA with the DINOSAur model instead of SLATE, and evaluation of ROCA on tasks with low-dimensional vector states are presented in the appendix.

\section{Conclusion and Future Work}
 We presented ROCA, an object-centric off-policy value-based model-based reinforcement learning approach that uses a pre-trained SLATE model as an object-centric feature extractor.
 Our experiments in 3D and 2D tasks demonstrate that ROCA learns effective policies and outperforms object-centric model-free and model-based baselines.
 The world model is built upon a GNN architecture, showing that graph neural networks can be successfully applied in MBRL settings for policy learning.
 While we use the SLATE model as an object-centric feature extractor, in principle, we can replace SLATE with other slot-based object-centric models.
 However, ROCA does have limitations.
 Firstly, its world model is deterministic and may struggle to predict the dynamics of highly stochastic environments.
 Additionally, as our model is based on the SAC algorithm, it is sensitive to the target entropy hyperparameter, especially in environments with discrete action spaces~\citep{xu2021target, zhou2023revisiting}.

 In our future work, we consider the primary task to be evaluating ROCA in more visually challenging environments.
 To accomplish this, we plan to replace SLATE with the recently proposed DLP~\citep{daniel2024ddlpunsupervisedobjectcentricvideo} model, which has shown promising results in goal-conditioned tasks when paired with the model-free algorithm~\citep{haramati2024entitycentricreinforcementlearningobject}.

\section*{Acknowledgements}
This work was supported by Russian Science Foundation, grant No. 20-71-10116, \url{https://rscf.ru/en/project/20-71-10116}. The research was carried out using the infrastructure of the Shared Research Facilities "High Performance Computing and Big Data" (CKP "Informatics") of FRC CSC RAS (Moscow).

\bibliography{bibliography}

\begin{thebibliography}{54}
\providecommand{\natexlab}[1]{#1}
\providecommand{\url}[1]{\texttt{#1}}
\expandafter\ifx\csname urlstyle\endcsname\relax
  \providecommand{\doi}[1]{doi: #1}\else
  \providecommand{\doi}{doi: \begingroup \urlstyle{rm}\Url}\fi

\bibitem[Ahmed et~al.(2020)Ahmed, Träuble, Goyal, Neitz, Bengio, Schölkopf,
  Wüthrich, and Bauer]{ahmed2020causalworld}
Ossama Ahmed, Frederik Träuble, Anirudh Goyal, Alexander Neitz, Yoshua Bengio,
  Bernhard Schölkopf, Manuel Wüthrich, and Stefan Bauer.
\newblock Causalworld: A robotic manipulation benchmark for causal structure
  and transfer learning, 2020.

\bibitem[Biza et~al.(2022)Biza, Platt, van~de Meent, Wong, and
  Kipf]{biza2022binding}
Ondrej Biza, Robert Platt, Jan-Willem van~de Meent, Lawson~L.S. Wong, and
  Thomas Kipf.
\newblock Binding actions to objects in world models.
\newblock In \emph{ICLR2022 Workshop on the Elements of Reasoning: Objects,
  Structure and Causality}, 2022.
\newblock URL \url{https://openreview.net/forum?id=HImz8BuUclc}.

\bibitem[Burgess et~al.(2019)Burgess, Matthey, Watters, Kabra, Higgins,
  Botvinick, and Lerchner]{burgess2019monet}
Christopher~P. Burgess, Loic Matthey, Nicholas Watters, Rishabh Kabra, Irina
  Higgins, Matt Botvinick, and Alexander Lerchner.
\newblock Monet: Unsupervised scene decomposition and representation, 2019.

\bibitem[Christodoulou(2019)]{christodoulou2019soft}
Petros Christodoulou.
\newblock Soft actor-critic for discrete action settings, 2019.

\bibitem[Daniel and Tamar(2024)]{daniel2024ddlpunsupervisedobjectcentricvideo}
Tal Daniel and Aviv Tamar.
\newblock Ddlp: Unsupervised object-centric video prediction with deep dynamic
  latent particles, 2024.
\newblock URL \url{https://arxiv.org/abs/2306.05957}.

\bibitem[Diuk et~al.(2008)Diuk, Cohen, and Littman]{10.1145/1390156.1390187}
Carlos Diuk, Andre Cohen, and Michael~L. Littman.
\newblock An object-oriented representation for efficient reinforcement
  learning.
\newblock In \emph{Proceedings of the 25th International Conference on Machine
  Learning}, ICML '08, page 240–247, New York, NY, USA, 2008. Association for
  Computing Machinery.
\newblock ISBN 9781605582054.
\newblock \doi{10.1145/1390156.1390187}.
\newblock URL \url{https://doi.org/10.1145/1390156.1390187}.

\bibitem[Engelcke et~al.(2021)Engelcke, Parker~Jones, and
  Posner]{NEURIPS2021_43ec517d}
Martin Engelcke, Oiwi Parker~Jones, and Ingmar Posner.
\newblock Genesis-v2: Inferring unordered object representations without
  iterative refinement.
\newblock In M.~Ranzato, A.~Beygelzimer, Y.~Dauphin, P.S. Liang, and J.~Wortman
  Vaughan, editors, \emph{Advances in Neural Information Processing Systems},
  volume~34, pages 8085--8094. Curran Associates, Inc., 2021.
\newblock URL
  \url{https://proceedings.neurips.cc/paper_files/paper/2021/file/43ec517d68b6edd3015b3edc9a11367b-Paper.pdf}.

\bibitem[Engelcke et~al.(2022)Engelcke, Jones, and
  Posner]{engelcke2022genesisv2}
Martin Engelcke, Oiwi~Parker Jones, and Ingmar Posner.
\newblock Genesis-v2: Inferring unordered object representations without
  iterative refinement, 2022.

\bibitem[Farquhar et~al.(2018)Farquhar, Rocktäschel, Igl, and
  Whiteson]{farquhar2018treeqn}
Gregory Farquhar, Tim Rocktäschel, Maximilian Igl, and Shimon Whiteson.
\newblock Treeqn and atreec: Differentiable tree-structured models for deep
  reinforcement learning, 2018.

\bibitem[Ferraro et~al.(2023)Ferraro, Mazzaglia, Verbelen, and
  Dhoedt]{ferraro2023focusobjectcentricworldmodels}
Stefano Ferraro, Pietro Mazzaglia, Tim Verbelen, and Bart Dhoedt.
\newblock Focus: Object-centric world models for robotics manipulation, 2023.
\newblock URL \url{https://arxiv.org/abs/2307.02427}.

\bibitem[Fujimoto et~al.(2018)Fujimoto, van Hoof, and
  Meger]{fujimoto2018addressingfunctionapproximationerror}
Scott Fujimoto, Herke van Hoof, and David Meger.
\newblock Addressing function approximation error in actor-critic methods,
  2018.
\newblock URL \url{https://arxiv.org/abs/1802.09477}.

\bibitem[Funk et~al.(2022)Funk, Chalvatzaki, Belousov, and
  Peters]{pmlr-v164-funk22a}
Niklas Funk, Georgia Chalvatzaki, Boris Belousov, and Jan Peters.
\newblock Learn2assemble with structured representations and search for robotic
  architectural construction.
\newblock In Aleksandra Faust, David Hsu, and Gerhard Neumann, editors,
  \emph{Proceedings of the 5th Conference on Robot Learning}, volume 164 of
  \emph{Proceedings of Machine Learning Research}, pages 1401--1411. PMLR,
  08--11 Nov 2022.
\newblock URL \url{https://proceedings.mlr.press/v164/funk22a.html}.

\bibitem[Haarnoja et~al.(2018)Haarnoja, Zhou, Abbeel, and
  Levine]{haarnoja2018soft}
Tuomas Haarnoja, Aurick Zhou, Pieter Abbeel, and Sergey Levine.
\newblock Soft actor-critic: Off-policy maximum entropy deep reinforcement
  learning with a stochastic actor, 2018.

\bibitem[Haarnoja et~al.(2019)Haarnoja, Zhou, Hartikainen, Tucker, Ha, Tan,
  Kumar, Zhu, Gupta, Abbeel, and Levine]{haarnoja2019soft}
Tuomas Haarnoja, Aurick Zhou, Kristian Hartikainen, George Tucker, Sehoon Ha,
  Jie Tan, Vikash Kumar, Henry Zhu, Abhishek Gupta, Pieter Abbeel, and Sergey
  Levine.
\newblock Soft actor-critic algorithms and applications, 2019.

\bibitem[Hafner et~al.(2022)Hafner, Lillicrap, Norouzi, and
  Ba]{hafner2022masteringataridiscreteworld}
Danijar Hafner, Timothy Lillicrap, Mohammad Norouzi, and Jimmy Ba.
\newblock Mastering atari with discrete world models, 2022.
\newblock URL \url{https://arxiv.org/abs/2010.02193}.

\bibitem[Hafner et~al.(2023)Hafner, Pasukonis, Ba, and
  Lillicrap]{hafner2023mastering}
Danijar Hafner, Jurgis Pasukonis, Jimmy Ba, and Timothy Lillicrap.
\newblock Mastering diverse domains through world models, 2023.

\bibitem[Haramati et~al.(2024)Haramati, Daniel, and
  Tamar]{haramati2024entitycentricreinforcementlearningobject}
Dan Haramati, Tal Daniel, and Aviv Tamar.
\newblock Entity-centric reinforcement learning for object manipulation from
  pixels, 2024.
\newblock URL \url{https://arxiv.org/abs/2404.01220}.

\bibitem[Jiang et~al.(2020)Jiang, Janghorbani, de~Melo, and
  Ahn]{jiang2020scalorgenerativeworldmodels}
Jindong Jiang, Sepehr Janghorbani, Gerard de~Melo, and Sungjin Ahn.
\newblock Scalor: Generative world models with scalable object representations,
  2020.
\newblock URL \url{https://arxiv.org/abs/1910.02384}.

\bibitem[Jocher et~al.(2022)Jocher, {Ayush Chaurasia}, Stoken, Borovec,
  {NanoCode012}, {Yonghye Kwon}, {Kalen Michael}, {TaoXie}, {Jiacong Fang},
  {Imyhxy}, {, Lorna}, {(Zeng Yifu)}, Wong, {Abhiram V}, Montes, {Zhiqiang
  Wang}, Fati, {Jebastin Nadar}, {Laughing}, {UnglvKitDe}, Sonck, {Tkianai},
  {YxNONG}, Skalski, Hogan, {Dhruv Nair}, Strobel, and
  Jain]{https://doi.org/10.5281/zenodo.3908559}
Glenn Jocher, {Ayush Chaurasia}, Alex Stoken, Jirka Borovec, {NanoCode012},
  {Yonghye Kwon}, {Kalen Michael}, {TaoXie}, {Jiacong Fang}, {Imyhxy}, {,
  Lorna}, {(Zeng Yifu)}, Colin Wong, {Abhiram V}, Diego Montes, {Zhiqiang
  Wang}, Cristi Fati, {Jebastin Nadar}, {Laughing}, {UnglvKitDe}, Victor Sonck,
  {Tkianai}, {YxNONG}, Piotr Skalski, Adam Hogan, {Dhruv Nair}, Max Strobel,
  and Mrinal Jain.
\newblock ultralytics/yolov5: v7.0 - yolov5 sota realtime instance
  segmentation, 2022.
\newblock URL \url{https://zenodo.org/record/3908559}.

\bibitem[Keramati et~al.(2018)Keramati, Whang, Cho, and
  Brunskill]{keramati2018fast}
Ramtin Keramati, Jay Whang, Patrick Cho, and Emma Brunskill.
\newblock Fast exploration with simplified models and approximately optimistic
  planning in model based reinforcement learning, 2018.

\bibitem[Kipf et~al.(2020)Kipf, van~der Pol, and Welling]{Kipf2020Contrastive}
Thomas Kipf, Elise van~der Pol, and Max Welling.
\newblock Contrastive learning of structured world models.
\newblock In \emph{International Conference on Learning Representations}, 2020.
\newblock URL \url{https://openreview.net/forum?id=H1gax6VtDB}.

\bibitem[Kirilenko et~al.(2023)Kirilenko, Korchemnyi, Smirnov, Kovalev, and
  Panov]{kirilenko_quantized_2023}
Daniil Kirilenko, Alexandr Korchemnyi, Konstantin Smirnov, Alexey~K Kovalev,
  and Aleksandr~I Panov.
\newblock Quantized disentangled representations for object-centric visual
  tasks.
\newblock In \emph{Pattern Recognition and Machine Intelligence. {PReMI} 2023.
  Lecture Notes in Computer Science}, volume 14301, pages 514--522. Springer
  Cham, 2023.
\newblock \doi{10.1007/978-3-031-45170-6_53}.
\newblock URL
  \url{https://link.springer.com/chapter/10.1007/978-3-031-45170-6_53}.

\bibitem[Kirilenko et~al.(2024)Kirilenko, Vorobyov, Kovalev, and
  Panov]{kirilenko_object-centric_2024}
Daniil Kirilenko, Vitaliy Vorobyov, Alexey Kovalev, and Aleksandr Panov.
\newblock Object-centric learning with slot mixture module.
\newblock In \emph{The Twelfth International Conference on Learning
  Representations}, 2024.
\newblock URL \url{https://openreview.net/forum?id=aBUidW4Nkd}.

\bibitem[Kirillov et~al.(2023)Kirillov, Mintun, Ravi, Mao, Rolland, Gustafson,
  Xiao, Whitehead, Berg, Lo, Dollár, and Girshick]{kirillov2023segment}
Alexander Kirillov, Eric Mintun, Nikhila Ravi, Hanzi Mao, Chloe Rolland, Laura
  Gustafson, Tete Xiao, Spencer Whitehead, Alexander~C. Berg, Wan-Yen Lo, Piotr
  Dollár, and Ross Girshick.
\newblock Segment anything, 2023.
\newblock URL \url{https://arxiv.org/abs/2304.02643}.

\bibitem[Lin et~al.(2020)Lin, Wu, Peri, Sun, Singh, Deng, Jiang, and
  Ahn]{lin2020spaceunsupervisedobjectorientedscene}
Zhixuan Lin, Yi-Fu Wu, Skand~Vishwanath Peri, Weihao Sun, Gautam Singh, Fei
  Deng, Jindong Jiang, and Sungjin Ahn.
\newblock Space: Unsupervised object-oriented scene representation via spatial
  attention and decomposition, 2020.
\newblock URL \url{https://arxiv.org/abs/2001.02407}.

\bibitem[Locatello et~al.(2020{\natexlab{a}})Locatello, Weissenborn,
  Unterthiner, Mahendran, Heigold, Uszkoreit, Dosovitskiy, and
  Kipf]{NEURIPS2020_8511df98}
Francesco Locatello, Dirk Weissenborn, Thomas Unterthiner, Aravindh Mahendran,
  Georg Heigold, Jakob Uszkoreit, Alexey Dosovitskiy, and Thomas Kipf.
\newblock Object-centric learning with slot attention.
\newblock In H.~Larochelle, M.~Ranzato, R.~Hadsell, M.F. Balcan, and H.~Lin,
  editors, \emph{Advances in Neural Information Processing Systems}, volume~33,
  pages 11525--11538. Curran Associates, Inc., 2020{\natexlab{a}}.
\newblock URL
  \url{https://proceedings.neurips.cc/paper_files/paper/2020/file/8511df98c02ab60aea1b2356c013bc0f-Paper.pdf}.

\bibitem[Locatello et~al.(2020{\natexlab{b}})Locatello, Weissenborn,
  Unterthiner, Mahendran, Heigold, Uszkoreit, Dosovitskiy, and
  Kipf]{locatello2020objectcentric}
Francesco Locatello, Dirk Weissenborn, Thomas Unterthiner, Aravindh Mahendran,
  Georg Heigold, Jakob Uszkoreit, Alexey Dosovitskiy, and Thomas Kipf.
\newblock Object-centric learning with slot attention, 2020{\natexlab{b}}.

\bibitem[Locatello et~al.(2020{\natexlab{c}})Locatello, Weissenborn,
  Unterthiner, Mahendran, Heigold, Uszkoreit, Dosovitskiy, and
  Kipf]{slotattention}
Francesco Locatello, Dirk Weissenborn, Thomas Unterthiner, Aravindh Mahendran,
  Georg Heigold, Jakob Uszkoreit, Alexey Dosovitskiy, and Thomas Kipf.
\newblock Object-centric learning with slot attention, 2020{\natexlab{c}}.

\bibitem[Mnih et~al.(2015)Mnih, Kavukcuoglu, Silver, Rusu, Veness, Bellemare,
  Graves, Riedmiller, Fidjeland, Ostrovski, Petersen, Beattie, Sadik,
  Antonoglou, King, Kumaran, Wierstra, Legg, and Hassabis]{Mnih2015}
Volodymyr Mnih, Koray Kavukcuoglu, David Silver, Andrei~A. Rusu, Joel Veness,
  Marc~G. Bellemare, Alex Graves, Martin Riedmiller, Andreas~K. Fidjeland,
  Georg Ostrovski, Stig Petersen, Charles Beattie, Amir Sadik, Ioannis
  Antonoglou, Helen King, Dharshan Kumaran, Daan Wierstra, Shane Legg, and
  Demis Hassabis.
\newblock Human-level control through deep reinforcement learning.
\newblock \emph{Nature}, 518\penalty0 (7540):\penalty0 529--533, February 2015.
\newblock \doi{10.1038/nature14236}.
\newblock URL \url{https://doi.org/10.1038/nature14236}.

\bibitem[Ramesh et~al.(2021)Ramesh, Pavlov, Goh, Gray, Voss, Radford, Chen, and
  Sutskever]{dalle}
Aditya Ramesh, Mikhail Pavlov, Gabriel Goh, Scott Gray, Chelsea Voss, Alec
  Radford, Mark Chen, and Ilya Sutskever.
\newblock Zero-shot text-to-image generation, 2021.

\bibitem[Santoro et~al.(2017)Santoro, Raposo, Barrett, Malinowski, Pascanu,
  Battaglia, and Lillicrap]{santoro2017simple}
Adam Santoro, David Raposo, David G.~T. Barrett, Mateusz Malinowski, Razvan
  Pascanu, Peter Battaglia, and Timothy Lillicrap.
\newblock A simple neural network module for relational reasoning, 2017.

\bibitem[Schulman et~al.(2017)Schulman, Wolski, Dhariwal, Radford, and
  Klimov]{https://doi.org/10.48550/arxiv.1707.06347}
John Schulman, Filip Wolski, Prafulla Dhariwal, Alec Radford, and Oleg Klimov.
\newblock Proximal policy optimization algorithms, 2017.
\newblock URL \url{https://arxiv.org/abs/1707.06347}.

\bibitem[Schölkopf et~al.(2021)Schölkopf, Locatello, Bauer, Ke, Kalchbrenner,
  Goyal, and Bengio]{9363924}
Bernhard Schölkopf, Francesco Locatello, Stefan Bauer, Nan~Rosemary Ke, Nal
  Kalchbrenner, Anirudh Goyal, and Yoshua Bengio.
\newblock Toward causal representation learning.
\newblock \emph{Proceedings of the IEEE}, 109\penalty0 (5):\penalty0 612--634,
  2021.
\newblock \doi{10.1109/JPROC.2021.3058954}.

\bibitem[Seitzer et~al.(2023)Seitzer, Horn, Zadaianchuk, Zietlow, Xiao,
  Simon-Gabriel, He, Zhang, Schölkopf, Brox, and
  Locatello]{seitzer2023bridging}
Maximilian Seitzer, Max Horn, Andrii Zadaianchuk, Dominik Zietlow, Tianjun
  Xiao, Carl-Johann Simon-Gabriel, Tong He, Zheng Zhang, Bernhard Schölkopf,
  Thomas Brox, and Francesco Locatello.
\newblock Bridging the gap to real-world object-centric learning, 2023.

\bibitem[Sharma et~al.(2023)Sharma, Gupta, Faldu, Gupta, ., and
  Singla]{sharma2023objectcentric}
Vishal Sharma, Aniket Gupta, Prayushi Faldu, Rushil Gupta, Mausam ., and Parag
  Singla.
\newblock Object-centric learning of neural policies for zero-shot transfer
  over domains with varying quantities of interest.
\newblock In \emph{PRL Workshop Series {\textendash} Bridging the Gap Between
  AI Planning and Reinforcement Learning}, 2023.
\newblock URL \url{https://openreview.net/forum?id=mQtyk75pYZ}.

\bibitem[Singh et~al.(2022)Singh, Deng, and Ahn]{singh2022illiterate}
Gautam Singh, Fei Deng, and Sungjin Ahn.
\newblock Illiterate dall-e learns to compose.
\newblock In \emph{ICLR}, 2022.

\bibitem[Stanić et~al.(2022)Stanić, Tang, Ha, and
  Schmidhuber]{stanić2022learning}
Aleksandar Stanić, Yujin Tang, David Ha, and Jürgen Schmidhuber.
\newblock Learning to generalize with object-centric agents in the open world
  survival game crafter, 2022.

\bibitem[Sutton and Barto(2018)]{10.5555/3312046}
Richard~S. Sutton and Andrew~G. Barto.
\newblock \emph{Reinforcement Learning: An Introduction}.
\newblock A Bradford Book, Cambridge, MA, USA, 2018.
\newblock ISBN 0262039249.

\bibitem[Szot et~al.(2022)Szot, Clegg, Undersander, Wijmans, Zhao, Turner,
  Maestre, Mukadam, Chaplot, Maksymets, Gokaslan, Vondrus, Dharur, Meier,
  Galuba, Chang, Kira, Koltun, Malik, Savva, and Batra]{szot2022habitat}
Andrew Szot, Alex Clegg, Eric Undersander, Erik Wijmans, Yili Zhao, John
  Turner, Noah Maestre, Mustafa Mukadam, Devendra Chaplot, Oleksandr Maksymets,
  Aaron Gokaslan, Vladimir Vondrus, Sameer Dharur, Franziska Meier, Wojciech
  Galuba, Angel Chang, Zsolt Kira, Vladlen Koltun, Jitendra Malik, Manolis
  Savva, and Dhruv Batra.
\newblock Habitat 2.0: Training home assistants to rearrange their habitat,
  2022.

\bibitem[Tao et~al.(2024)Tao, Xiang, Shukla, Qin, Hinrichsen, Yuan, Bao, Lin,
  Liu, kai Chan, Gao, Li, Mu, Xiao, Gurha, Huang, Calandra, Chen, Luo, and
  Su]{tao2024maniskill3gpuparallelizedrobotics}
Stone Tao, Fanbo Xiang, Arth Shukla, Yuzhe Qin, Xander Hinrichsen, Xiaodi Yuan,
  Chen Bao, Xinsong Lin, Yulin Liu, Tse kai Chan, Yuan Gao, Xuanlin Li,
  Tongzhou Mu, Nan Xiao, Arnav Gurha, Zhiao Huang, Roberto Calandra, Rui Chen,
  Shan Luo, and Hao Su.
\newblock Maniskill3: Gpu parallelized robotics simulation and rendering for
  generalizable embodied ai, 2024.
\newblock URL \url{https://arxiv.org/abs/2410.00425}.

\bibitem[van~den Oord et~al.(2018)van~den Oord, Vinyals, and
  Kavukcuoglu]{oord2018neural}
Aaron van~den Oord, Oriol Vinyals, and Koray Kavukcuoglu.
\newblock Neural discrete representation learning, 2018.

\bibitem[Vaswani et~al.(2023)Vaswani, Shazeer, Parmar, Uszkoreit, Jones, Gomez,
  Kaiser, and Polosukhin]{vaswani2023attention}
Ashish Vaswani, Noam Shazeer, Niki Parmar, Jakob Uszkoreit, Llion Jones,
  Aidan~N. Gomez, Lukasz Kaiser, and Illia Polosukhin.
\newblock Attention is all you need, 2023.

\bibitem[Watters et~al.(2019)Watters, Matthey, Bosnjak, Burgess, and
  Lerchner]{watters2019cobra}
Nicholas Watters, Loic Matthey, Matko Bosnjak, Christopher~P. Burgess, and
  Alexander Lerchner.
\newblock Cobra: Data-efficient model-based rl through unsupervised object
  discovery and curiosity-driven exploration, 2019.

\bibitem[Wu et~al.(2023)Wu, Dvornik, Greff, Kipf, and Garg]{wu2023slotformer}
Ziyi Wu, Nikita Dvornik, Klaus Greff, Thomas Kipf, and Animesh Garg.
\newblock Slotformer: Unsupervised visual dynamics simulation with
  object-centric models.
\newblock In \emph{The Eleventh International Conference on Learning
  Representations}, 2023.
\newblock URL \url{https://openreview.net/forum?id=TFbwV6I0VLg}.

\bibitem[Xu et~al.(2021)Xu, Hu, Liang, McAleer, Abbeel, and Fox]{xu2021target}
Yaosheng Xu, Dailin Hu, Litian Liang, Stephen McAleer, Pieter Abbeel, and Roy
  Fox.
\newblock Target entropy annealing for discrete soft actor-critic, 2021.

\bibitem[Ye et~al.(2021)Ye, Liu, Kurutach, Abbeel, and Gao]{ye2021mastering}
Weirui Ye, Shaohuai Liu, Thanard Kurutach, Pieter Abbeel, and Yang Gao.
\newblock Mastering atari games with limited data, 2021.

\bibitem[Yi et~al.(2022)Yi, Zhang, Peng, Guo, Hu, Du, Zhang, Guo, and
  Chen]{yi2022objectcategoryawarereinforcementlearning}
Qi~Yi, Rui Zhang, Shaohui Peng, Jiaming Guo, Xing Hu, Zidong Du, Xishan Zhang,
  Qi~Guo, and Yunji Chen.
\newblock Object-category aware reinforcement learning, 2022.
\newblock URL \url{https://arxiv.org/abs/2210.07802}.

\bibitem[Yoon et~al.(2023)Yoon, Wu, Bae, and Ahn]{yoon2023investigation}
Jaesik Yoon, Yi-Fu Wu, Heechul Bae, and Sungjin Ahn.
\newblock An investigation into pre-training object-centric representations for
  reinforcement learning, 2023.

\bibitem[Zadaianchuk et~al.(2020)Zadaianchuk, Seitzer, and
  Martius]{zadaianchuk2020selfsupervisedvisualreinforcementlearning}
Andrii Zadaianchuk, Maximilian Seitzer, and Georg Martius.
\newblock Self-supervised visual reinforcement learning with object-centric
  representations, 2020.
\newblock URL \url{https://arxiv.org/abs/2011.14381}.

\bibitem[Zadaianchuk et~al.(2022{\natexlab{a}})Zadaianchuk, Martius, and
  Yang]{zadaianchuk2022selfsupervised}
Andrii Zadaianchuk, Georg Martius, and Fanny Yang.
\newblock Self-supervised reinforcement learning with independently
  controllable subgoals, 2022{\natexlab{a}}.

\bibitem[Zadaianchuk et~al.(2022{\natexlab{b}})Zadaianchuk, Martius, and
  Yang]{zadaianchuk2022selfsupervisedreinforcementlearningindependently}
Andrii Zadaianchuk, Georg Martius, and Fanny Yang.
\newblock Self-supervised reinforcement learning with independently
  controllable subgoals, 2022{\natexlab{b}}.
\newblock URL \url{https://arxiv.org/abs/2109.04150}.

\bibitem[Zhang et~al.(2021)Zhang, McAllister, Calandra, Gal, and
  Levine]{zhang2021learning}
Amy Zhang, Rowan~Thomas McAllister, Roberto Calandra, Yarin Gal, and Sergey
  Levine.
\newblock Learning invariant representations for reinforcement learning without
  reconstruction.
\newblock In \emph{International Conference on Learning Representations}, 2021.
\newblock URL \url{https://openreview.net/forum?id=-2FCwDKRREu}.

\bibitem[Zholus et~al.(2022)Zholus, Ivchenkov, and Panov]{Zholus2022a}
Artem Zholus, Yaroslav Ivchenkov, and Aleksandr Panov.
\newblock Factorized {World} {Models} for {Learning} {Causal} {Relationships}.
\newblock In \emph{{ICLR} {Workshop} on the {Elements} of {Reasoning}:
  {Objects}, {Structure} and {Causality}}, 2022.
\newblock URL \url{https://openreview.net/forum?id=BCGfDBOIcec}.

\bibitem[Zhou et~al.(2023)Zhou, Lin, Li, Fu, Yang, and Ye]{zhou2023revisiting}
Haibin Zhou, Zichuan Lin, Junyou Li, Qiang Fu, Wei Yang, and Deheng Ye.
\newblock Revisiting discrete soft actor-critic, 2023.

\end{thebibliography}
\appendix

\section{Environments}

\subsection{Shapes2D}
\begin{itemize}
    \item \textbf{Navigation 5x5}: Number of objects: 5; Observation size: 50x50; Number of actions: 16; Episode length: 100.
    \item \textbf{PushingNoAgent 5x5}: Number of objects: 5; Observation size: 50x50; Number of actions: 16; Episode length: 100.
    \item \textbf{Navigation 10x10}: Number of objects: 8; Observation size: 100x100; Number of actions: 28; Episode length: 100.
\end{itemize}

\subsection{CausalWorld}
\begin{itemize}
    \item \textbf{Object Reaching}: Observation size: 64x64; Action dimensionality: 3; Episode length: 100.
\end{itemize}

\section{Object-Centric Representation Models}

\subsection{Datasets}

\paragraph{Shapes2D}
Training datasets of 300K observations are collected using a uniform random policy for every task, with the exception of Navigation 5x5 and PushingNoAgent 5x5. Since these tasks share the same observation space, we use only the Navigation 5x5 dataset.

\paragraph{CausalWorld}
For the Object Reaching task with the VOYC color scheme, we collect following the procedure described in the official OCRL repository. For the BGYR color scheme, we do not collect data as we use the pretrained SLATE model provided by the authors.

\subsection{SLATE}
\label{appendix:slate}
\begin{table*}[ht]
\centering
\begin{tabular}{cccc}
\midrule
& & Shapes2D & CausalWorld \\
\midrule
\multirow{10}{*}{Learning}
& Training dataset size & 300000 & 1000000 \\
& Temp. Cooldown & \multicolumn{2}{c}{1.0 to 0.1} \\
& Temp. Cooldown Steps & \multicolumn{2}{c}{30000} \\
& LR for DVAE & \multicolumn{2}{c}{0.0003} \\
& LR for CNN Encoder & \multicolumn{2}{c}{0.0001} \\
& LR for Transformer Decoder & \multicolumn{2}{c}{0.0003} \\
& LR Warm Up Steps & \multicolumn{2}{c}{30000} \\
& LR Half Time & \multicolumn{2}{c}{250000} \\
& Dropout & \multicolumn{2}{c}{0.1} \\
& Clip & \multicolumn{2}{c}{0.05} \\
& Batch Size & \multicolumn{2}{c}{32} \\
& Epochs & \multicolumn{2}{c}{100} \\
\midrule
DVAE & Vocabulary Size & 32 & 4096 \\
\midrule
CNN Encoder & Hidden Size & \multicolumn{2}{c}{64} \\
\midrule
\multirow{5}{*}{Slot Attention} & Iterations & 5 & 3\\
& Slot Heads & \multicolumn{2}{c}{1} \\
& Slot Dim. & 64 & 192\\
& MLP Hidden Dim. & 256 & 192 \\
\midrule
\multirow{3}{*}{Transformer Decoder} & Layers & \multicolumn{2}{c}{4} \\
& Heads & \multicolumn{2}{c}{4} \\
& Hidden Dim. & 128 & 192 \\
\midrule
\end{tabular}
\caption{Common Hyperparameters for SLATE}
\label{table:slate_hyperparams}
\end{table*}

\begin{table*}[ht]
\centering
\begin{tabular}{cccc}
Task & Resize & Num. Slots & Pos Channels \\
\midrule
Navigation 5x5 & 64 & 6 & 5 \\
PushingNoAgent 5x5 & 64 & 6 & 5 \\
Navigation 10x10 & 96 & 9 & 8 \\
Object Reaching & 64 & 10 & 4 \\
\midrule
\end{tabular}
\caption{Task-Specific Hyperparameters for SLATE}
\label{table:slate_task_hyperparams}
\end{table*}

Our code is based on the OCRL repository: \url{https://github.com/jsikyoon/OCRL}.

\noindent Hyperparameters for SLATE are listed in Tables \ref{table:slate_hyperparams} and \ref{table:slate_task_hyperparams}. For the Object Reaching task with the BGYR color scheme, we use model weights from the original OCRL repository.

 \begin{figure*}[ht]
   \centering
   \includegraphics[width=\textwidth]{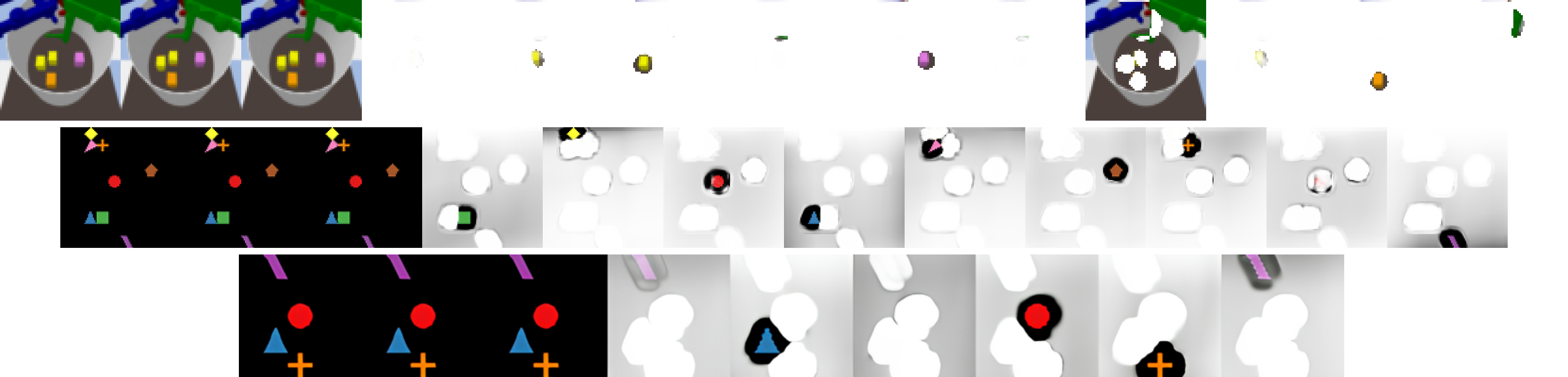}
   \caption{Visualization of attention maps produced by SLATE in Object Reaching and Shapes2D tasks} \label{fig:appendix_slots}
 \end{figure*}

\section{ROCA}
\label{appendix:roca}
 \begin{figure*}[ht]
   \centering
   \includegraphics[width=\textwidth]{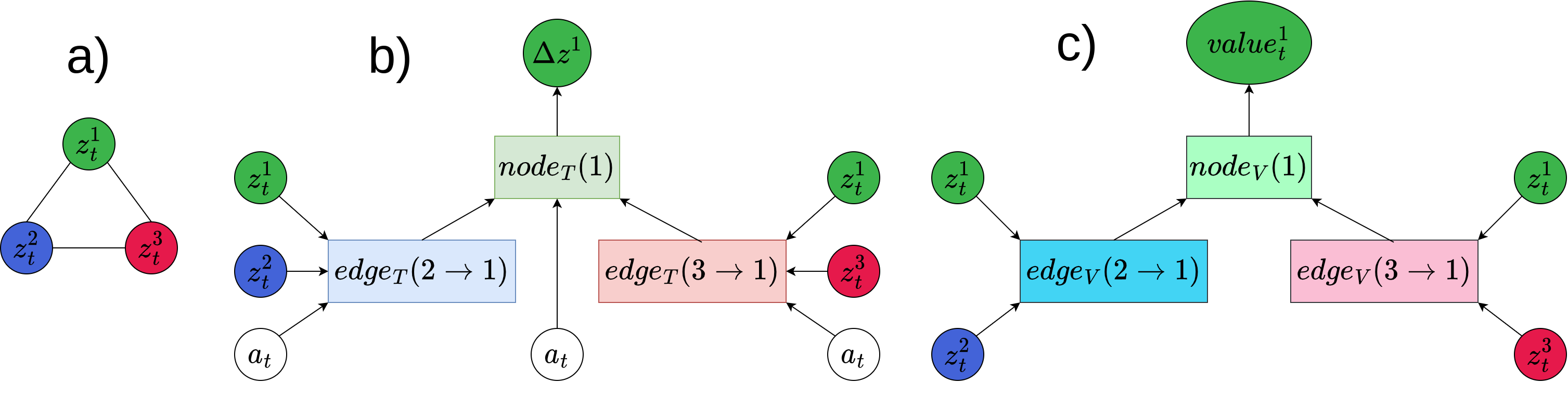}
   \caption{Overview of GNN-based transition model and state-value model. \textbf{a)} Representation of the state as a complete graph. \textbf{b)} Transition model: message-passing update scheme for the embedding of object 1. \textbf{c)} State-value model: message-passing update scheme for the state-value prediction for the object 1.} \label{fig:gnn}
 \end{figure*}

\begin{table*}[h]
\centering
\begin{tabular}{cc}
\midrule
Gamma & 0.99 \\
Buffer size & 1000000 \\
Batch size & 128 \\
$\tau_{polyak}$ & 0.005 \\
$\beta_{T}$ & 1 \\
$\beta_{R}$ & 1 \\
Buffer prefill size & 5000 \\
Number of parallel environments & 16 \\
Learning rate & 0.0003\\
Train frequency & Every step in environment \\
Target network update frequency & Every training step \\
\midrule
\end{tabular}
\caption{Hyperparameters for ROCA}
\label{table:ROCA_hyperparams}
\end{table*}
 
\texttt{edge} and \texttt{node} models in the GNN-based Transition model, Reward model, and Actor model are MLPs consisting of three hidden layers with 512 units each, along with LayerNorm and ReLU activations. The MLPs following the concatenation-readout layer in the Actor, Critic, Reward Predictor, and Continue Predictor models consist of a single layer with 512 units.

The target entropy parameter is set to -3 for the Object Reaching task. For tasks with a discrete action space, we scale the entropy of a uniform random policy with a coefficient of 0.6. This results in values of 1.66 for the Navigation 5x5 and PushingNoAgent 5x5 tasks, and 2 for the Navigation 10x10 task. The other hyperparameters are listed in Table \ref{table:ROCA_hyperparams}.

\noindent Tested hyperparameters:
\begin{itemize}
    \item Target entropy (Object Reaching): [-1, -2, -3, -4, -5]
    \item Target entropy (Shapes2D) --- scaling coefficients for the entropy of a uniform random distribution: [0.4, 0.5, 0.6, 0.7, 0.8, 0.9]
    \item Batch size: [64, 128, 256]
    \item Target network update frequency (every N training steps): [1, 2, 3]
\end{itemize}

\section{ROCA-CSWM}
\label{appendix:roca-cswm}
We use a CSWM-like encoder (https://github.com/tkipf/c-swm), which consists of four convolutional layers with BatchNorm layers and ReLU activations between them:
\begin{itemize}
    \item kernel\_size: 3, stride: 1, padding: 1, out\_channels: 32
    \item kernel\_size: 3, stride: 1, padding: 1, out\_channels: 32
    \item kernel\_size: 5, stride: 2, padding: 1, out\_channels: 32
    \item kernel\_size: 5, stride: 2, padding: 1, out\_channels: \texttt{num\_slots}
\end{itemize}
The value of the \texttt{num\_slots} parameter depends on the task: Navigation 5x5 and PushingNoAgent 5x5 --- 5, Navigation 10x10 --- 8, Object Reaching --- 6.

The output of the final convolutional layer is flattened and fed into an MLP consisting of three layers with ReLU activations between them. This MLP is shared between slots. Before the final layer, we use layer normalization. We use 128 hidden units in the MLP. The output dimension of the final layer in the MLP is the same as the Slot Dim. parameter in Table \ref{table:slate_hyperparams}. The other hyperparameters of ROCA-CSWM are listed in Table \ref{table:ROCA-CSWM_hyperparams}.

\begin{table*}[h]
\centering
\begin{tabular}{cc}
\midrule
Gamma & 0.99 \\
Buffer size & 1000000 \\
Batch size & 256 \\
$\tau_{polyak}$ & 0.005 \\
$\beta_{T}$ & 1 \\
$\beta_{R}$ & 1 \\
$\beta_{C}$ & 1 \\
$\gamma_{margin}$ & 0.025 \\
Buffer prefill size & 5000 \\
Number of parallel environments & 16 \\
Learning rate & 0.0003\\
Train frequency & Every step in environment \\
Target network update frequency & Every training step \\
\midrule
\end{tabular}
\caption{Hyperparameters for ROCA-CSWM}
\label{table:ROCA-CSWM_hyperparams}
\end{table*}

\noindent Tested hyperparameters:
\begin{itemize}
    \item Target entropy (Object Reaching): [-1, -2, -3, -4, -5]
    \item Target entropy (Shapes2D): Scaling coefficients for the entropy of a uniform random distribution: [0.4, 0.5, 0.6, 0.7, 0.8, 0.9]
    \item Batch size: [64, 128, 256, 512]
    \item Target network update frequency (every N training steps): [1, 2, 3]
    \item $\gamma_{margin}$: [5, 1, 0.5, 0.025, 0.01]
\end{itemize}

\section{DreamerV3, DreamerV3:pretrained}
For DreamerV3, we use the \texttt{medium} preset for configuration parameters with the \texttt{train\_ratio} parameter increased to 512. We also ran experiments with the \texttt{large} and \texttt{xlarge} presets but did not notice any improvements. We resize observations to the same sizes as OCR models (Table \ref{table:slate_task_hyperparams}).

\noindent Our code is based on the original DreamerV3 repository: \url{https://github.com/danijar/dreamerv3}

\section{DreamerV3:slate}
Similarly, for DreamerV3:slate, we use the \texttt{medium} preset because the \texttt{large} and \texttt{xlarge} presets did not produce performance gains. We also set the \texttt{train\_ratio} parameter to 512. The input of DreamerV3:slate is the concatenation of slots, which are vectors of size $\texttt{num\_slots} \times \texttt{slot\_dim}$ (Tables \ref{table:slate_hyperparams} and \ref{table:slate_task_hyperparams}).

\noindent Our code is based on the original DreamerV3 repository: \url{https://github.com/danijar/dreamerv3}

\section{OCRL}
Our code is based on the OCRL repository: \url{https://github.com/jsikyoon/OCRL}

\noindent PPO-related tested hyperparameters:
\begin{itemize}
    \item Entropy coefficient: [0, 0.001, 0.01, 0.025, 0.05, 0.075, 0.1]
    \item Clip range: [0.1, 0.2, 0.4]
    \item Epochs: [10, 20, 30]
    \item Batch size: [32, 64, 128]
    \item GAE lambda: [0.90, 0.95]
\end{itemize}

The best hyperparameters we found are:
\begin{itemize}
    \item Entropy coefficient: 0.01
    \item Clip range: 0.2
    \item Epochs: 10
    \item Batch size: 32
    \item GAE lambda: 0.95
\end{itemize}

\section{Ablation Study}
We perform a grid search to find the best hyperparameters for the baselines. We use the same parameter ranges as those used for ROCA.

\section{Additional Experiments with DreamerV3}
\label{appendix:dreamer}
\begin{figure*}[!htb]
  \centering
  \includegraphics[width=0.98\textwidth]{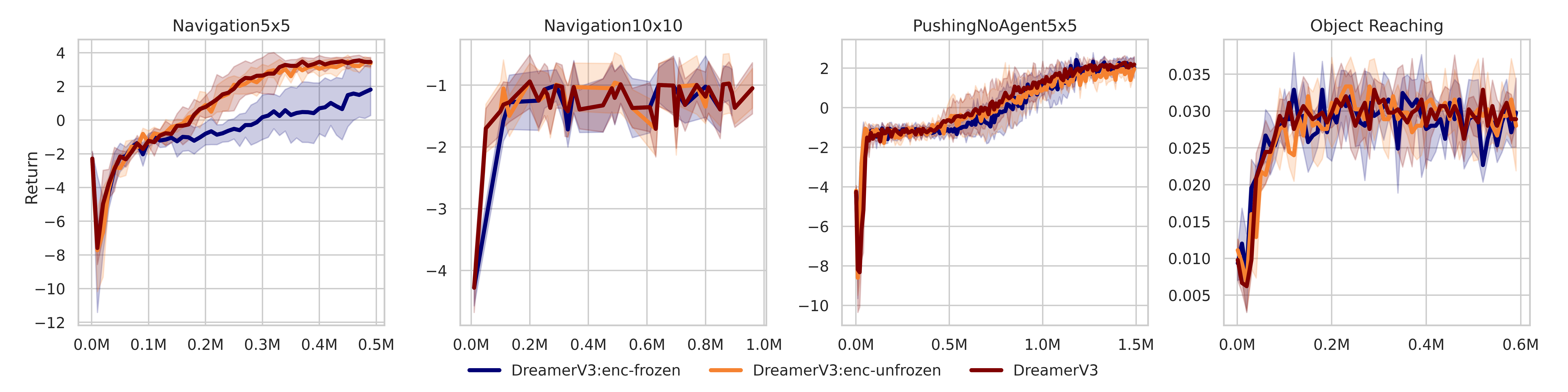}
  \caption{The plots illustrate the impact of encoder pretraining on the DreamerV3 algorithm. \textit{DreamerV3} is the default version that trains its encoder from scratch. \textit{DreamerV3:enc-frozen} is a version with a pretrained frozen encoder. \textit{DreamerV3:enc-unfrozen} is a version with a pretrained unfrozen encoder. Returns are averaged over 30 episodes and three seeds.} 
  \label{fig:dreamer_pretrained}
\end{figure*}

To ensure a fair comparison with DreamerV3, we conducted experiments using a pretrained encoder obtained from the DreamerV3 model that solves the task. For all tasks, we tested two different modes: one with the encoder frozen and another with the encoder unfrozen. However, we did not observe any improvement in the convergence rate. The results are shown in Figure \ref{fig:dreamer_pretrained}.

\section{Evaluation of Out-of-Distribution Generalization to Unseen Colors}
\label{appendix:ood}
We evaluated the generalization of ROCA to unseen colors of distractor objects in the Object Reaching task. When tested with the same colors it was trained on, the model achieved a success rate of $0.975 \pm 0.005$. However, when new colors were used for the distractor objects, the success rate dropped to $0.850 \pm 0.005$. The results were averaged over three instances of the ROCA model, with each instance evaluated over 100 episodes.

\section{Sparse Interactions Modeling}
\label{appendix:sparse}
\begin{figure*}[!htb]
  \centering
  \includegraphics[width=0.98\textwidth]{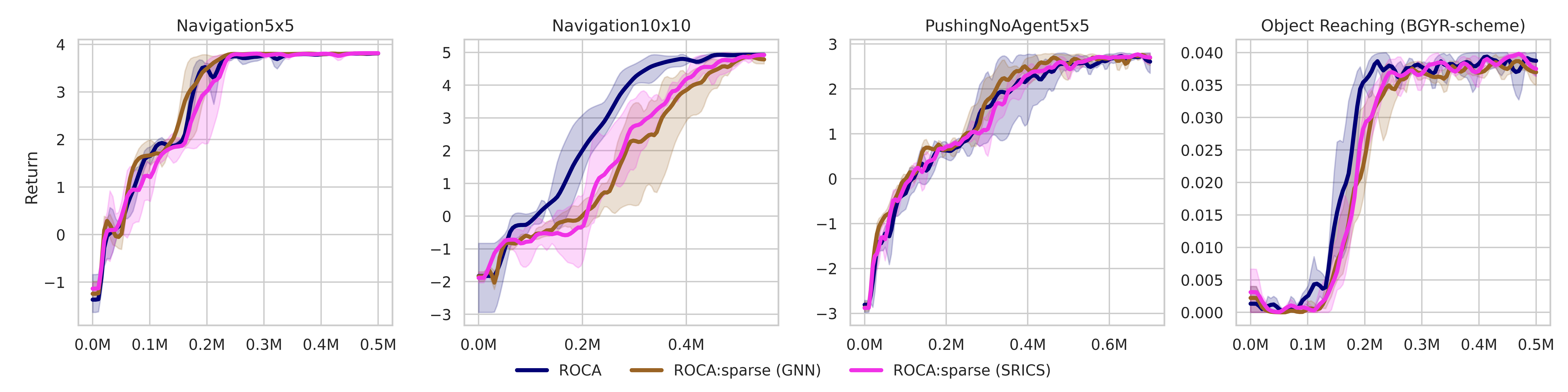}
  \caption{The plots illustrate the performance of variations of the ROCA model with explicit modeling of sparse interactions between objects. The results show that ROCA performs better than the modifications.}
  \label{fig:sparse}
\end{figure*}

In ROCA, we assume dense interactions among objects, modeled using complete graphs. However, interactions in real-world scenarios often occur sparsely rather than densely. While GNNs can emulate these naturally sparse interactions through their neural representations, some works explicitly model these interactions using a sparse graph~\cite{zadaianchuk2022selfsupervised}. To determine if modeling sparse interactions could improve our algorithm's efficiency, we implemented two modifications.

In \textbf{ROCA:sparse (GNN)}, we implemented an additional GNN-based $Interaction_{GNN}$ model. The \texttt{node} function of this model produces a weight $0 \leq w_t^i \leq 1$ for each object. This weight represents the degree of involvement of object $i$ at time step $t$ in interactions. These weights $w_t^i$ are used as multipliers for \texttt{edge} functions in the transition, reward, state-value, and actor models. For example, the modified Transition model is:
\begin{equation}
    \Delta z^i = \texttt{node}_T(z_t^i, a_t^i, \sum_{j \neq i} w_t^i w_t^j \texttt{edge}_T(z_t^i, z_t^j, a_t)).
\end{equation}
No additional loss function is used to train the $Interaction_{GNN}$ model.

In \textbf{ROCA:sparse (SRICS)}, we employ the approach proposed with the SRICS model~\cite{zadaianchuk2022selfsupervised}. We implement an $Interaction_{SRICS}$ model as an \texttt{edge} function. The output $w_t^{ij}$ of $Interaction_{SRICS}$ is treated as a parameter of a Bernoulli distribution. A value of one indicates an interaction between objects $i$ and $j$ at time step $t$, while a value of zero indicates no interaction. Similar to $Interaction_{GNN}$, we use $w_t^{ij}$ as a multiplier for \texttt{edge} functions in the transition, reward, state-value, and actor models. For example, the modified Transition model is:
\begin{equation}
    \Delta z^i = \texttt{node}_T(z_t^i, a_t^i, \sum_{j \neq i} w_t^{ij} \texttt{edge}_T(z_t^i, z_t^j, a_t)).
\end{equation}
To enforce sparsity, we use the Kullback-Leibler divergence $D_{KL}(w_t^{ij}, p_{prior})$ between the Bernoulli distribution induced by $Interaction_{SRICS}$ and the prior distribution $p_{prior}$:
\begin{equation}
    p_{prior}(k; p) = \begin{cases}
    p, & \text{if } k = 1, \\
    1 - p, & \text{if } k = 0.
    \end{cases}
\end{equation}
As in the original implementation~\cite{zadaianchuk2022selfsupervised}, we use $p = 0.05$, which means a low probability of interactions. We add the KL-divergence to the loss functions of the Critic, Actor, and World Model.

Figure \ref{fig:sparse} demonstrates the results of our experiments with the ROCA modifications. We did not observe any significant improvements from modeling sparse interactions.

\section{Performance of ROCA with DINOSAur Slot-Extractor Model}
\label{appendix: dinosaur}
 \begin{figure*}[ht]
   \centering
   \includegraphics[width=0.4 \textwidth]{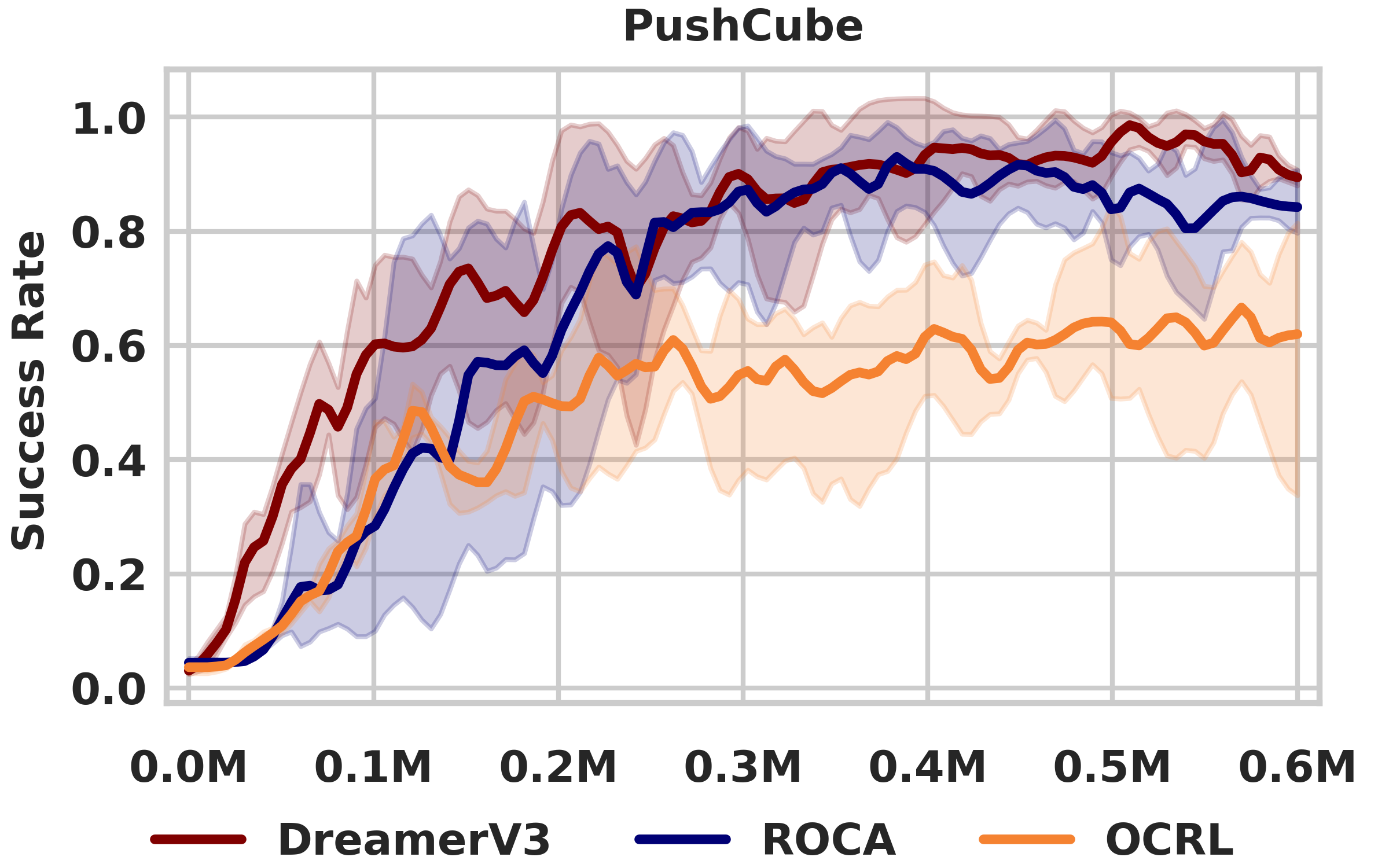}
   \caption{The plots illustrate the performance of ROCA and the baselines on the PushCube task in the Maniskill environment. Success rate is averaged over 30 episodes and three seeds.} \label{fig:appendix_maniskill}
 \end{figure*}
We additionally trained ROCA, OCRL, and DreamerV3 baselines on the PushCube task in the ManiSkill~\citep{tao2024maniskill3gpuparallelizedrobotics} environment.
In both OCRL and ROCA, we replaced SLATE with DINOSAur~\citep{seitzer2023bridging}, as it is expected to perform better in more visually complex domains.
In our experiments, we found that selecting four slots yielded the best average FG-ARI (foreground-adjusted Rand Index) value of 65\% ± 10\% on a validation set of 10K images (compared to three, five, and six slots).
To calculate FG-ARI, we used ground-truth four-class segmentation: background, robotic arm, target goal region, and cube.
The results are presented in Figure \ref{fig:appendix_maniskill}.
We found that ROCA outperforms OCRL but learns more slowly than DreamerV3.
We attribute this to the lower quality of object-centric representations produced by DINOSAur in the PushCube task, compared to SLATE in the Shapes2D environment, where FG-ARI scores exceed 90\%.

\section{Performance of ROCA with Low-dimensional Vector States}
\label{appendix: state}
 \begin{figure*}[ht]
   \centering
   \includegraphics[width= \textwidth]{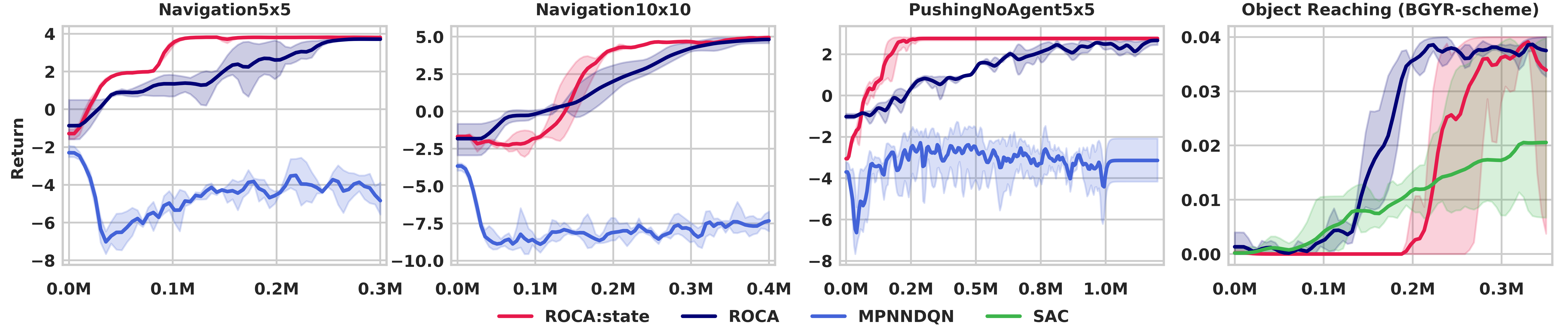}
   \caption{The plots illustrate the performance of ROCA and models that use low-dimensional vector states as input: MPNNDQN and ROCA:state. Since MPNNDQN cannot be used in tasks with continuous action spaces, we replace it with SAC for the Object Reaching task. The return is averaged over 30 episodes and three seeds.} \label{fig:appendix_state}
 \end{figure*}
We conducted experiments on variants of the Shapes2D and Object Reaching tasks, where ROCA is trained on low-dimensional states of objects. We compare the performance of the standard ROCA algorithm (as presented in this paper) with ROCA:state (where vector states of objects are used as input instead of slots) and MPNNDQN~\citep{pmlr-v164-funk22a}—a modification of DQN~\citep{Mnih2015} based on a multi-head attention GNN.

MPNNDQN is built on top of DQN, and because of this, it cannot be applied to tasks with continuous action spaces, such as Object Reaching. The original architecture of MPNNDQN is tightly coupled to the environment, as described in the paper. It assumes that the environment provides information about which vertices are adjacent in the graph representation of the scene. However, our environments do not provide this information, so in our adaptation, we use complete graphs. Additionally, MPNNDQN was designed for action spaces with a hierarchical structure, where actions are defined relative to pairs of objects. As a result, the original MPNNDQN produces sets of action values for every pair of objects. In contrast, our environments require action values for the entire scene, so we add a non-learned readout layer that pools these values over all pairs by summation. The results, presented in Figure \ref{fig:appendix_state}, show that ROCA:state outperforms the standard ROCA on the Shapes2D tasks. In contrast, ROCA converges faster than ROCA:state on the Object Reaching task. For our version of MPNNDQN, we performed a limited hyperparameter search, but the algorithm still underperforms. We attribute this behavior to the specific use case for which MPNNDQN was originally designed. For completeness, we also include results from Soft Actor-Critic (SAC) on the Object Reaching task, demonstrating that ROCA:state outperforms SAC

\section{Compute Resources}
\label{appendix:computer_resources}
The experiments were carried out on a cluster with eight Nvidia Tesla V100 GPUs (32 GB each), 48 Intel(R) Xeon(R) Gold 2.60 GHz CPU cores, and 1536 GB RAM. We also used a cluster with three Nvidia Titan RTX GPUs (24 GB each), 80 Intel(R) Xeon(R) Gold 2.60 GHz CPU cores, and 1024 GB RAM. Additionally, we occasionally used a desktop computer with an Nvidia RTX 2070 GPU, 6 Intel(R) Core i5-9600K CPUs, and 32 GB RAM. The rough estimate of the total computational cost required to reproduce the experiments is approximately 13,000 CPU-hours and 5,000 GPU-hours.

\end{document}